\documentclass{tADR2e}
\usepackage{ulem}
\usepackage{color}    
\usepackage{jumoline} 
\usepackage{proofread}%

\noproofreadmark

\usepackage{empheq}

\DeclareMathOperator*{\argmax}{argmax}
\DeclareMathOperator*{\argmin}{argmin}

\usepackage{algorithm}
\usepackage{algorithmic}

\begin{document}
    \jvol{xx} \jnum{xx} \jyear{xxxx} \jmonth{xxxxxxx}
    \title{Bayesian Body Schema Estimation\\ using Tactile Information obtained through \addspan{Coordinated} Random Movements}
    \author{Tomohiro Mimura$^{a \ast}$\thanks{$^\ast$. Email: t.mimura@em.ci.ritsumei.ac.jp\vspace{6pt}},
        Yoshinobu Hagiwara$^{a}$, Tadahiro Taniguchi$^{a}$ and Tetsunari Inamura$^{b,c}$\\
        \vspace{6pt}
        $^{a}${\em{Ritsumeikan University \\ 1-1-1 Noji Higashi, Kusatsu, Shiga 525-8577, Japan}}\\
        $^{b}${\em{National Institute of Informatics \\2-1-2 Hitotsubashi, Chiyoda-ku, Tokyo 101-8430, Japan}}\\
        $^{c}${\em{The Graduate University for Advanced Studies \\2-1-2 Hitotsubashi, Chiyoda-ku, Tokyo 101-8430, Japan}}\\
        \vspace{6pt}\received{v1.0 released January 2017}}
    \maketitle
    \begin{abstract}
        \addspan{This paper describes a computational model, called the Dirichlet process Gaussian mixture model with latent joints (DPGMM-LJ), that can find latent tree structure embedded in data distribution in an unsupervised manner. By combining DPGMM-LJ and a pre-existing body map formation method, we propose a method that enables an agent having multi-link body structure to discover its kinematic structure, i.e.,  body schema, from tactile information alone.  The DPGMM-LJ is a probabilistic model based on Bayesian nonparametrics and an extension of Dirichlet process Gaussian mixture model (DPGMM).
            In a simulation experiment, we used a simple fetus model that had five body parts and performed  structured random movements in a womb-like environment. It was shown that the method could estimate the number of body parts and kinematic structures without any pre-existing knowledge in many cases. 
            Another experiment showed that the degree of motor coordination in random movements affects the result of body schema formation strongly.  It is confirmed that the accuracy rate for body schema estimation had the highest value $84.6$\% when the ratio of motor coordination was 0.9 in our setting.  These results suggest that kinematic structure can be estimated from tactile information obtained by a fetus moving randomly in a womb without any visual information even though its accuracy was not so high.  They also suggest that a certain degree of motor coordination in random movements and the sufficient dimension of state space that represents the body map are important to estimate body schema correctly. }
            \delspan{\\      
            This paper describes a computational model that enables an agent, a model of a fetus, to discover
its body schema from tactile information obtained by tactile sensors placed on the surface of every
body part through general movements. This is a constructive model of a fetus forming its body
schema through general movements. General movements performed by a fetus in its mother's womb
are spontaneous, random, and structured. In this study, we assume that a body schema means a
kinematic structure of a body system represented by a tree structure, i.e., a graph without loops. To
develop the computational model, we propose an unsupervised machine learning method called the
Dirichlet process Gaussian mixture model with latent joints (DPGMM-LJ) that can estimate clusters
of data points, the number of clusters, joint points of each cluster, and its latent tree structure,
simultaneously. The DPGMM-LJ is a probabilistic model based on Bayesian nonparametrics, and its
capability of estimating the latent structure of data points 
exibly is attributable to the nature of
Bayesian nonparametrics. First, the computational model forms a body map, i.e., a topological map of
the tactile sensors, from the time-series data of tactile information obtained while an agent performs
general movements. Second, it estimates clusters of sensors on the body map and a latent tree structure
representing the kinematic structure of the agent by inferring latent variables of the DPGMM-LJ. A
simulation experiment shows that the computational model can estimate the body schema, i.e., the
latent tree structure, of the agent correctly in many cases. Another experiment suggests that a certain
degree of motor coordination in general movements and the sucient dimension of representing the
body schema are important to estimate body schema correctly, even though estimating the number
of body parts hardly require such conditions.}
        \medskip
    \end{abstract}

    \section{Introduction}~\label{sec1}
    A human brain has a model of a body including a body map and a body schema~\cite{gallagher1986body,blakeslee2008body}. 
    A correct model of the body helps its owner to behave appropriately in his/her environment. However, once the model of the body becomes dysfunctional, his/her sensory-motor coordination, motor functions, and recognition of his/her posture become impaired~\cite{Ota2014}. 
    The question now arises: How can a human brain form a body schema?
    Particularly, this paper focuses on the body schema formation in the fetal period, and investigate the possible computational model that enables a fetus to form a body schema. 
    The goal of this study is to 
    develop a computational model that can enable an agent, a model of a fetus, to form its body schema described as a tree structure from tactile information obtained through \addspan{random}\delspan{general} movements just as a human fetus does.
    
    \begin{figure}[tb]
        \begin{center}
            \subfigure{
                \resizebox*{10cm}{!}{\includegraphics{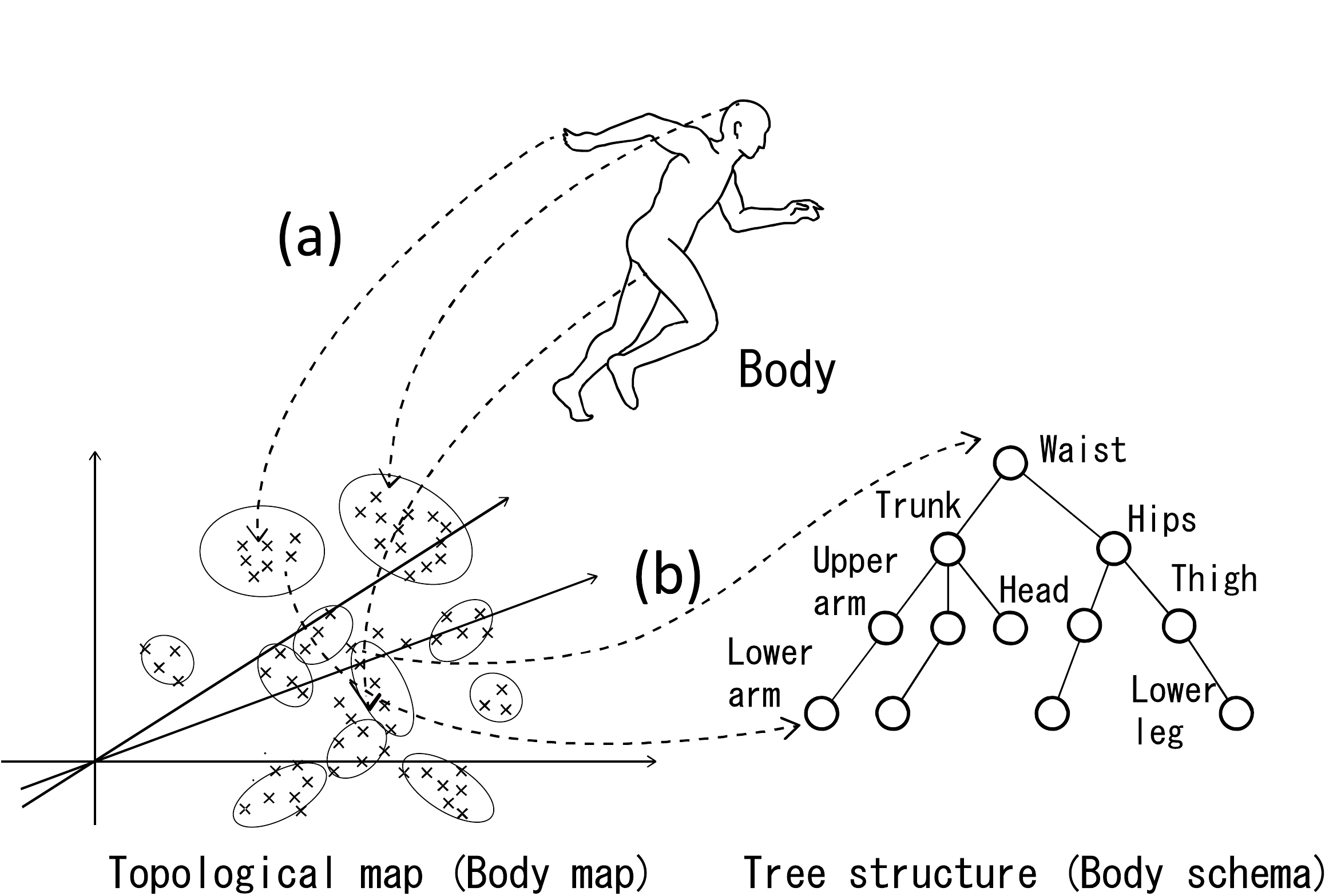}}}
            \caption{Overview of this research}
            \label{fig:overview}
        \end{center}
    \end{figure}
    
    From the computational point of view, the model of a body involves several components.
    For example, topological map, kinematic structure, physical parameters, kinematics, and dynamics are regarded as the components.
    Topological map, which is also called {\it body map}, holds information about positional and causal relationships between sensors and motors.
    A body map is a model of a human somatosensory map. Kinematic structure describes the number of body parts and their relations of connections. It is common that a human body system is modeled as a tree structure, i.e., a graph structure without any loops. This assumption is often used when we model kinematics and dynamics of humanoid robots~\cite{handbook}. Physical parameters of the body include the length of body parts, e.g., upper and lower arms, and the weight of the parts. Kinematics and dynamics modeled inside of a brain are often referred 
    as internal models involving forward and inverse models~\cite{Wolpert1998}.
    The components are mutually related.
    For example, the models of kinematics and dynamics assumed a fixed body structure and parameters. Therefore, we should clarify which component we are focusing on and the contribution of the proposed computational model.
    In this study, we focus on the computational model representing a person's kinematic structure. The model is referred as a {\it body schema}.

\addspan{An intuitive survey about body representation for robots and animals is provided by 
        Hoffmann et al.~\cite{Hoffmann2010}. There are several reasons why a human brain should form an internal representation of the person's kinematic structure. One of the most important reasons is that a human brain has to deal with coordination transformation. When a person tries to reach and grasp an external object, e.g., an apple or a cup, he/she must transform positional information of the target object to appropriate posture information.
        In other words, visual information should be transformed to posture information of its arm to move its end-effector's position to the target object, appropriately. This process is often modeled by using state space models. In such a case, each dimension of a state space, i.e., state variable, usually represents a variable of each joint, e.g., joint angle and angular velocity. However, this means that a brain cannot form an appropriate state space without knowledge of its kinematic structure including the number of body parts. The formation of a kinematic structure is a prerequisite for learning and adaptation based on state space model.    
    }
    
    It is highly probable that body schema is gradually formed during the fetal period to some extent.  
    Motor development during a fetal period is important.
    General movements are characteristic whole-body spontaneous movements performed by fetuses and infants~\cite{Einspieler2005,Einspieler1997,Prechtl1997,Hadders-Algra2004,Cioni1997}. Prechtl et al. conducted a variety of studies about general movements and proposed a diagnostic tool for early detection of brain dysfunction~\cite{Einspieler2005,Einspieler1997,Prechtl1997}. 
    Taga et al. analyzed the trajectories of hands and feet in general movements using a nonlinear prediction method~\cite{Taga1999}.
    Mori et al. developed a simulation environment of fetus exhibiting general movements and a computational model of its nervous system~\cite{Mori2010}.
    However, a computational model that can estimate the kinematic structure of the fetus's body has not been proposed. This paper describes a new model that can determine the number of body parts and the body's tree structure automatically only from tactile information that can be obtained while an agent representing a fetus conducts \addspan{random}\delspan{general} movements in its environment filled with liquid representing mother's womb, i.e., a virtual womb.
    
    To construct a computational model of body schema formation, the nonparametric Bayesian approach is promising.
    When a method attempts to estimate body schema, the method has to determine the number of body parts, and the latent tree structure. Nonparametric Bayesian approach enables a probabilistic model to determine the number of clusters and the other latent structure automatically from data. 
    \addspan{
        Conventionally, model selection methods using model selection criteria, e.g., AIC, BIC, and MDL, have been used to let a machine learner to optimize its model structure, e.g., the number of clusters and the network connections. However, to evaluate a possible configuration, e.g., the number of clusters, a whole learning process involving many iterations is required. When we try to use such a model selection methods to estimate the number of a body part and the body structure, the optimization process inevitably includes the combinatorial optimization process. This means the computational complexity becomes unrealistically huge. The nonparametric Bayesian approach can search for appropriate configuration of the model configuration through a single learning process in a stochastic manner. It can dramatically reduce the computational cost to estimate the latent tree structure, i.e., the body schema, efficiently.       
    }
    Therefore, we take a nonparametric Bayesian approach to constructing a computational model of body schema formation.

    The main contributions of this paper are as follows:
    \begin{enumerate}
        \item We propose Dirichlet process Gaussian mixture models with latent joints (DPGMM-LJ) that can cluster the data points, determine the number of Gaussian components, and 
        estimate its latent tree structure by inferring latent joint points in the state space. 
        \item We invent a new computational model that can estimate the body structure of an agent solely from its tactile information that can be obtained through \addspan{random}\delspan{general} movements in a virtual womb-like environment by combining a topological map formation method and the DPGMM-LJ. The computational model is evaluated through a simulation experiment.   
        \item Based on the constructive experiment, we obtained a result suggesting that a sufficient dimension of latent space for body map and a certain amount of motor coordination are important for forming an appropriate body schema, i.e., estimating kinematic structure, though they are not strongly required for estimating the number of body parts. 
    \end{enumerate}
    The overview of our proposed computational model is illustrated in Figure~\ref{fig:overview}. We assume that a human fetus forms a {\it body map}, and estimates a {\it body schema} using the formed body map.
    In our constructive model, an agent representing a human fetus performs \delspan{general}\addspan{random} movements in a liquid representing the amniotic fluid in a mother's womb, and obtains tactile information continuously. From whole-body tactile information, the agent estimates relative placement of the sensors in its latent state space in a bottom-up manner (see Figure~\ref{fig:overview}~(a)). The sensors placed in the latent state space are regarded as a body map estimated by the agent. 
    The body map formation method used in this paper is based on the information theory-based method proposed by Olsson et al.~\cite{olsson2006unknown}.  
    Next, the agent infers the latent tree structure, i.e., a body schema, by estimating clusters of sensors and joint points simultaneously (see Figure~\ref{fig:overview}~(b)). The simultaneous estimation of sensor clusters representing body parts, e.g., lower and upper arms, and the joints that connect two different body parts is conducted by the DPGMM-LJ.
    In this paper, we newly propose the DPGMM-LJ for modeling the process of body schema estimation.
    The DPGMM-LJ is a nonparametric Bayesian method that can estimate the number of body parts and joints solely from data and can infer the latent link structure embedded in 
    data distribution.

    The remainder of this paper is organized as follows. 
    Section~\ref{sec2} describes the background and related works of this study.
    Section~\ref{sec3} presents the DPGMM-LJ.  
    Section~\ref{sec4} describes our total constructive model that can estimate body schema from tactile information obtained through general movements-like exploratory random movements.
    Section~\ref{sec5} evaluates the effectiveness of the proposed method using simulation environment. Section~\ref{sec6} presents another experiment to investigate the relationship between the structure of \addspan{random}\delspan{general} movements and body schema formation.
    Section~\ref{sec7} concludes this paper.
    \section{Background}~\label{sec2}
    To clarify the background of our study, we provide a short survey of the previous studies about body map formation and body schema estimation.
    \subsection{Body map formation from tactile sensory information flow}
    Human body system has many body parts, and each body part has several tactile sensors.
    To determine the position where a tactile information is obtained, the human brain system should know the relationship between tactile sensors and body parts and the relative positions of the sensors on the body parts. It is believed that the human brain has a topological map that models allocation of tactile sensors on
    his/her body parts~\cite{kuniyoshi2004humanoid}． 
    The topological map of a human body in the brain is often called a body map~\cite{blakeslee2008body}.
    A body map covers various body parts, e.g., fingers, shoulders, and elbows, like an exhaustive patchwork.

    Several constructive studies have dealt with the problem of formation of a body map by a robot.  
    Kuniyoshi et al. proposed a constructive model that generates a topographic somatosensory cluster map that plots the fetus's sensor points on a two-dimensional space as a result of a self-organizing process ~\cite{kuniyoshi2004humanoid}.
    Olson et al. proposed a computational model that can reconstruct the layout of sensors and actuators by using only an information flow obtained from unknown sensors and actuators~\cite{olsson2006unknown}. Their method could successfully reconstruct a map of sensors and actuators of a robot without any human intervention. 
    The method uses distance metrics based on information theory. 
    Sugiura et al. solved a similar problem by using the time difference of arrival of signals and multidimensional scaling~\cite{Sugiura2006}. Yoshikawa et al. proposed a cross-modal map learning method for body schema acquisition~\cite{Yoshikawa2002}.
    
    However, these studies have been constructing models that can estimate only topological maps, i.e., body maps.
    The estimated body maps do not have information about the kinematic structure of an agent. 
    A body schema representing kinematic structure as a multilink system is essential for an agent to control its body. Kinematics and dynamics are generally formulated by using an assumed multilink system. 
    
    This means that the human brain system cannot control the body appropriately without generating its body schema. However, a computational model that can estimate body schema from a body map formed through \addspan{random}\delspan{general} movements is yet to be developed.
    This paper describes a novel computational model that can form body schema from tactile sensory information flow.
    For body map formation, our proposed method uses a model obtained by modifying the model proposed by Olsson et al.~\cite{olsson2006unknown}.

    \subsection{Body schema formation based on general movements}
    Here, we are concerned with the question: How can the brain system of a person discover his/her body structure, i.e., body schema, from his/her tactile sensory information? This is a long-standing problem. 
    Originally, Henry et al. introduced the term body schema when he discussed disordered spatial representation of the body following parietal lobe damage~\cite{head1911sensoey}. Even though the term body schema involves a variety of topics, from the computational viewpoint, in this paper we limit the discussion to the estimation of a multilink kinematic structure of a body system.
    Particularly, we focus on the body schema formation process by a fetus in its mother’s womb.

    It is highly probable that human fetuses start forming models of their bodies using their general movements that they perform inside of their mothers' wombs to a certain extent.
    Mori et al. provided a constructive model of a human fetus that involves a precise and complex body model of the musculoskeletal system and a model of neurodynamics~\cite{Mori2010}. They showed that the model of the fetus forms structured behaviors. They clearly modeled the self-organization process of the neural system of a fetus who obtains tactile information while moving in the womb, by measuring the resistance of the liquid---amniotic fluid---surrounding him/her. Our study is strongly inspired by this study and follows the assumption and experimental setting even though our experimental setting is simpler than that of Mori et al., and concentrates on computationally forming a body schema.  
    However, their model did not deal with the explicit estimation of body schema, i.e., kinematic structure, but the model of body obtained in the fetus's brain system is expressed only implicitly. Our study provides a computational model that can explicitly determine an agent's kinematic structure from whole-body tactile information obtained through \addspan{random}\delspan{general} movements.
    
    Developing a method that enables a robot to discover its body schema is regarded as a constructive approach towards modeling a human brain system that can form a body schema. 
    They assumed that the robot could observe the position of its end-effector. 
    Nabeshima et al. proposed a constructive model that enables a robot to modify its body schema based on multimodal information when robots use tools~\citep{Nabeshima2006}. This study is a constructive approach toward the modification of body schema in tool-use that was observed in Iriki et al.'s experiment~\cite{Iriki1996}.
    However, their studies did not treat the problem of kinematic structure estimation.

    In robotics, few studies about the estimation of the body structure of a robot have been conducted. Sturm et al. proposed a method that enables a robot to estimate its kinematic structure in an unsupervised manner using self-perception achieved by using a camera that observes the robot's posture from outside of the robot~\cite{Sturm2008}. Martinez et al. proposed a learning method that enables a robot to estimate the parameters of its body schema---, i.e., a kinematic model---using an active-learning method based on vision information~\cite{martinez2010body}.
    In computer vision, kinematic structures of a human body and other multilink movable objects are important to track them and estimate their postures. Several studies have explored methods that can extract kinematic structure from video data or motion capture data~\cite{Meeds2008,Chang2015,Chang2016,Fayad2011,Ross2010}.
    For example, Meads et al. proposed a nonparametric Bayesian model that can estimate stick-figure models that are regarded as body schemata~\citep{Meeds2008}.
    By virtue of the Bayesian nonparametrics, the method can automatically estimate the number of body parts and their joints by inferring the latent variables of the model.
    Out proposed method, the DPGMM-LJ can also automatically estimate the number of body parts and their joints by virtue of the Bayesian nonparametrics.
    Chang et al. proposed a method that can estimate the kinematic structure of articulated objects from a single-view image sequence~\cite{Chang2015}.
    However, all these studies assumed that the system could obtain visual information of the robot from a camera placed outside of the robot's embodied system.
    Hence, they cannot be regarded as a constructive model of the body schema formation process by the brain of a fetus.

    Based on the previous studies, we propose a computational model that enables an agent having a multilink body structure to estimate body schema, i.e., the number of body parts, correspondence between tactile sensors and body parts, and kinematic structure, in an unsupervised manner using tactile information obtained through \addspan{random}\delspan{general} movements by the agent.
    \section{Dirichlet Process Gaussian Mixture Model with Latent Joints (DPGMM-LJ)}~\label{sec3}
    \addspan{Our original contribution is to develop a new probabilistic graphical model DPGMM-LJ by extending DPGMM and deriving its inference procedure. 
        DPGMM-LJ is a probabilistic generative model that can be used to estimate not only the number of clusters but also latent tree structure from observed data points through a single clustering process. This can be used for estimating the kinematic structure and can estimate the number of Body parts by body map.}
    This section describes our proposed machine learning method based on a probabilistic generative model and Bayesian inference.
    The DPGMM-LJ proposed can cluster data points, determine the number of cluster components, and estimate the latent tree structure by inferring joint points of clusters, simultaneously.
    \addspan{
        Differently from kinematic structure estimation from visual images (e.g.,~\cite{Chang2016}), it is difficult to introduce pre-existing knowledge for clustering tactile information to estimate kinematic structure. Therefore, we simply introduce Gaussian distribution as elemental distribution for probabilistic clustering task and develop a general clustering method that can simultaneously estimate the latent link structure. 
    } Our constructive model attempts to extract kinematic structure information by applying the DPGMM-LJ to a formed body map (see Section \ref{subsec3-2}).
    
    \subsection{Dirichlet Process Gaussian Mixture Model (DPGMM)\label{DPGMM_model}}~\label{sec3_1}
    \begin{figure}[tb]
        \begin{center}
            \includegraphics[width=0.7\linewidth]{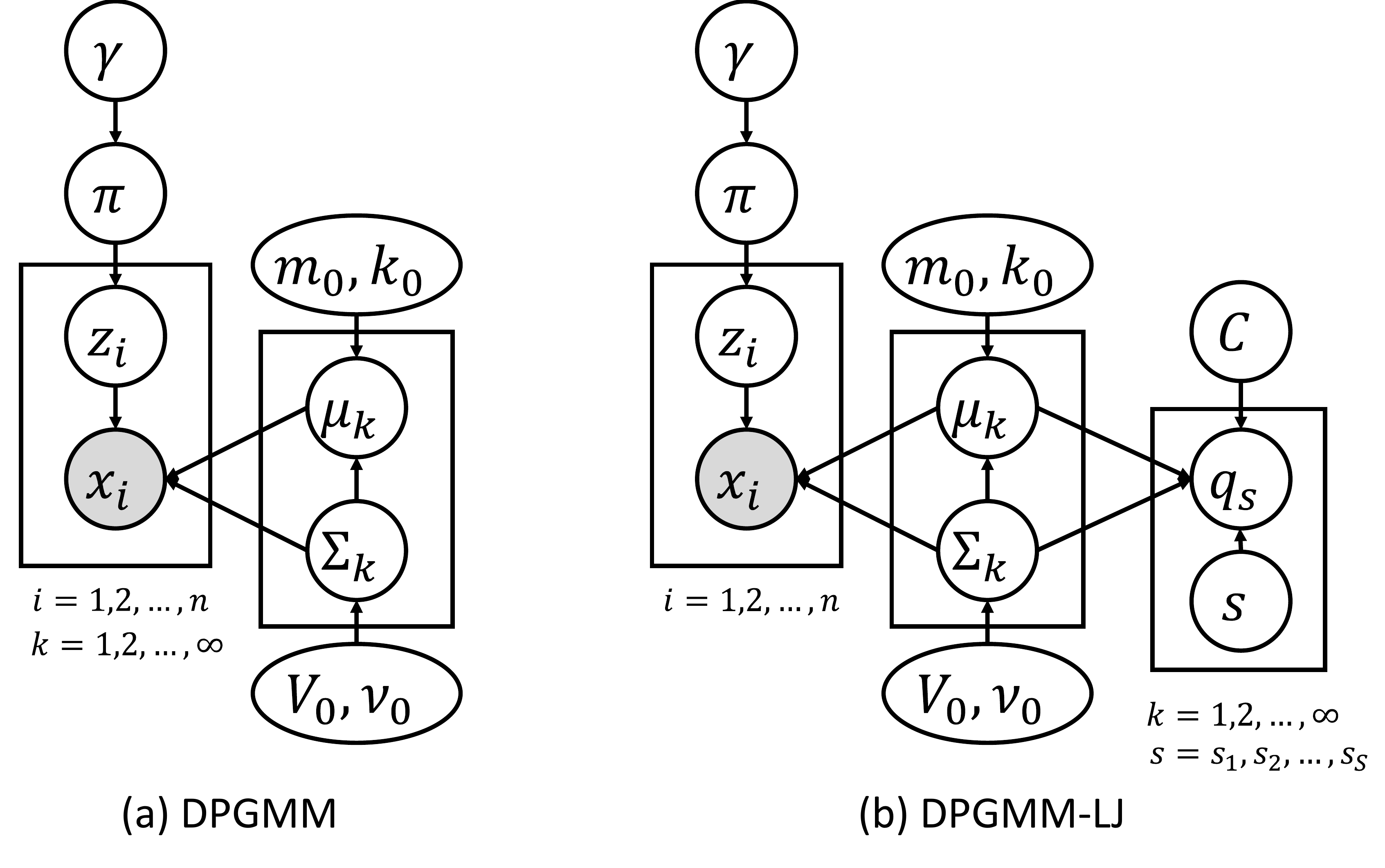}
            \caption{Graphical model of DPGMM and DPGMM-LJ}
            \label{fig:gm}
        \end{center}
    \end{figure}
    \addspan{A Gaussian mixture model (GMM) is a probabilistic model that assumes all the data points are generated from a mixture of Gaussian distributions, and it has been widely used for clustering problems.　Generally, a GMM has a fixed number of Gaussian distributions, i.e., the number of clusters, in advance. A GMM cannot infer its number of elemental distributions through its inference process. The DPGMM is a nonparametric Bayesian extension of the GMM, and a type of Dirichlet process mixtures (DPMs)~\cite{Blei2006}. Dirichlet Process  (DP) is a prior distribution over infinite dimensional multinomial distribution. It means that DP theoretically assumes an infinite number of clusters in advance. However, practically, a finite number of clusters are used for modeling observed data points. As a result, a DPGMM can automatically determine the number of clusters through its inference process in contrast with that the GMM cannot determine the number of clusters for itself~\cite{bishop2006pattern}. Figure~\ref{fig:gm} (a) shows a graphical model of the DPGMM representing the generative process of data points.  The graphical model expresses the conditional dependence structure between random variables on a probabilistic model.}  
    
    \addspan{The probability of generating  a data point by a DPGMM is calculated from the following equations:
        \begin{eqnarray}
        \pi &\sim &{\rm DP}(\gamma), \label{pimm_DPGMM} \\ 
        z_{i} &\sim& {\rm Categorical}(z|\pi), \label{Z_imm_DPGMM}\\
        \Sigma_{k}^{-1} &\sim&  {\cal W}(\Lambda | V_{0},\nu_{0}), \label{sigmamm_DPGMM}\\
        \mu_{k} &\sim& {\cal N}(\mu | m_{0},(k_{0} \Lambda)^{-1}), \label{mu_kmm_DPGMM}\\
        x_{i} &\sim& {\cal N}(x| {\mu}_{z_{i}} , \Sigma_{z_{i}}),  \label{x_imm_DPGMM}
        \end{eqnarray}
        where $\Lambda$ is a precision matrix calculated as an inverse matrix of a variance $\Sigma$. ${\rm Categorical}(\cdot)$ and ${\cal N}(\cdot)$ are a categorical distribution and a multivariate Gaussian distribution, respectively.  ${\rm DP}(\cdot)$ of Equation (\ref{pimm_DPGMM}) is Dirichlet process that generated a mixing ratio of the infinite number of Gaussian distribution　 $\pi$.} 
    
    \begin{table}[tb]
        \begin{center}
            \caption{Variables of DPGMM in Figure~\ref{fig:gm}}
            \begin{tabular}{ | c | c |} \hline
                Variables & Explanation\\\hline  \hline
                $x_{i} \in \mathbb{R}^d$  & \shortstack{The $i$-th data point.}\\   \hline 
                $\mu_{k}\in \mathbb{R}^d$, $\Sigma_{k} \in \mathbb{R}^{d\times d} $ & Mean and variance of the $k$-th Gaussian component \\ \hline 
                $\pi \in \mathbb{R}^\infty $ &  \shortstack{Parameter of the categorical distribution \\over the indexes of Gaussian components}\\ \hline 
                $z_{i} \in \mathbb{N}$ &Index of Gaussian distribution from which the $i$-th data point was sampled \\ \hline
                $\gamma$ & Hyperparameter of $\pi$ \\ \hline
                $m_{0},k_{0},V_{0},v_{0}$ & Hyperparameter of normal-Wishart distribution\\ 
                \hline
            \end{tabular}
            \label{tab:dpGMM}
        \end{center}
    \end{table}
    \addspan{
        Table \ref{tab:dpGMM} shows the variables of the graphical model in Figure~\ref{fig:gm} (a). Each of the random variables, i.e, $\pi, z, \mu, \Sigma$ and $x$, are generated by these equations from Equation (\ref{pimm_DPGMM}) to (\ref{x_imm_DPGMM}). The variable $x$ is representing observation. The another variables of DPGMM $m_{0}, k_{0}, V_{0}, v_{0}$ and $\gamma$ are hyperparameter that is parameter of prior distribution. DPGMM is defined by following equations:
        \begin{eqnarray}
        &p&(\bm{x_{1:n}}, \pi, z_{1:n}, \mu_{1:\infty}, \Sigma_{1:\infty}|\gamma,m_{0}, k_{0}, V_{0}, v_{0}) \label{DPGMM_equation} \nonumber \\
        &=& \prod_{k=1}^{\infty} \prod_{i=1}^{n} {\cal N}(x_i| {\mu}_{k} , \Sigma_{k}) {\rm Categorical}(z_i|\pi){\rm DP}(\gamma){\cal N}(\mu_k |m_{0},(k_{0} \Lambda_k)^{-1}){\cal W}(\Lambda_k | V_{0},\nu_{0}).
        \end{eqnarray}}
    
    \subsection{Generative model of proposed method (DPGMM-LJ)}
\begin{figure}[tb]
        \begin{center}
            \includegraphics[width=1.0\linewidth]{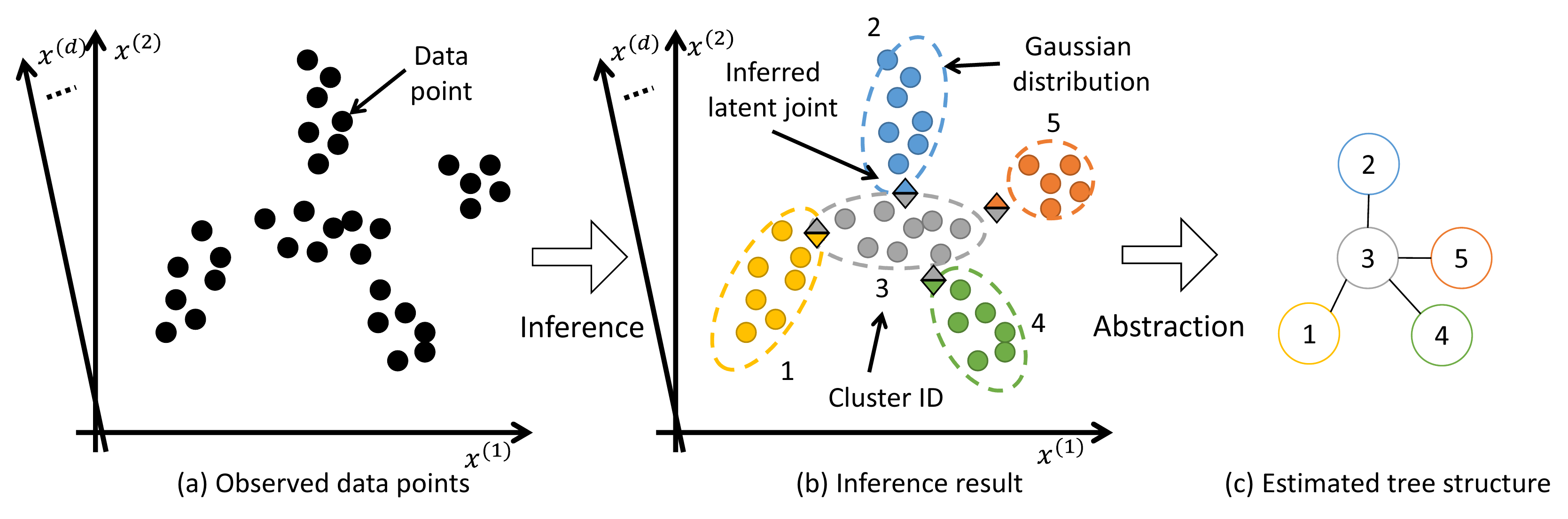}
            \caption{Overview of the inference process of DPGMM-LJ}
            \label{fig:dpGMM-LJ}
        \end{center}
    \end{figure}

    Figure~\ref{fig:dpGMM-LJ} shows the overview of the DPGMM-LJ and its inference process.
    First, we assume that a set of data points $x_i = (x^{(1)}_i, x^{(2)}_i, \ldots , x^{(d)}_i) \in \mathbb{R}^d ~  (i = 1, 2, \ldots , n)$ is observed (Figure~\ref{fig:dpGMM-LJ} (a)). Second, Bayesian inference is performed. Each dotted circle represents a cluster of data points. Each cluster is expected to correspond to a body part. 
    Moreover, between clusters, latent joint points are inferred so as to discover the tree structure embedded in the set of data points. 
    Third, by abstracting the relationship among clusters, the latent tree structure can be extracted. 
    \begin{table}[tb]
        \begin{center}
            \caption{Variables of DPGMM-LJ in Figure~\ref{fig:gm}}
            \begin{tabular}{ | c | c |} \hline
                Variables & Explanation\\\hline  \hline
                $x_{i} \in \mathbb{R}^d$ & \shortstack{The $i$-th data point, i.e., position of the $i$-th tactile sensor,\\ on a body map} \\   \hline 
                $\mu_{k}\in \mathbb{R}^d$, $\Sigma_{k} \in \mathbb{R}^{d\times d} $ & Mean and variance of the $k$-th Gaussian component \\ \hline 
                $\pi \in \mathbb{R}^\infty $ &  \shortstack{Parameter of the categorical distribution \\over the indexes of Gaussian components}\\ \hline 
                $z_{i} \in \mathbb{N}$ &Index of Gaussian distribution from which the $i$-th data point was sampled \\ \hline
                $q_{s}\in \mathbb{R}^d$  & Position of the $s$-th latent joint\\ \hline
                $C$  & \shortstack{Tree structure represented by a set of parent nodes for each node. \\ $C=\{c_k\}_{k\in \mathbb{N}}$ where $c_k \in \mathbb{N}\cup \{ \textrm{root-node} \}$. }  \\ \hline
                $\gamma$ & Hyperparameter of $\pi$ \\ \hline
                $m_{0},k_{0},V_{0},v_{0}$ & Hyperparameter of normal-Wishart distribution\\ 
                \hline
            \end{tabular}
            \label{tab:dpGMM-LJ}
        \end{center}
    \end{table}
    
    Figure~\ref{fig:gm} (b) shows a graphical model of the DPGMM-LJ representing the  generative process of data points having a latent tree structure. By inferring the latent variables of the model using data points on a body map, we can estimate a body schema.
    Table \ref{tab:dpGMM-LJ} shows the variables of the graphical model in Figure~\ref{fig:gm} (b).
    Graphical models represent the structure of conditional dependence between random variables. For simplicity, we use the notation $ (\cdot) = \{ (\cdot)_{k} \}_{\forall k}$, e.g., $\mu = \{ \mu_{k} \}_{\forall k}$ in the following sections.
    Our proposed model was constructed by extending the DPGMM that is one of the generative models based on Bayesian nonparametrics used for clustering (see Figure~\ref{fig:gm} (b)). 
    
    \delspan{The DPGMM is a nonparametric Bayesian extension of the GMM, and a type of Dirichlet process mixtures (DPMs)~\cite{Blei2006}. 
        The DPGMM can automatically determine the number of clusters through its inference process although the GMM cannot determine the number of clusters for itself~\cite{bishop2006pattern}.}
    In our problem settings, we assume that the learning system does not know the number of body parts, i.e., clusters of tactile sensors, before the learning phase.
    Specifically, we use the stick-breaking process (SBP) to construct the Dirichlet process~\cite{sethuraman1991constructive}.
    We employ weak-limit approximation for representing the SBP in a practical and approximate manner~\cite{HDP,Fox2009}. In the following sections, we describe the inference algorithm based on the assumption.
    
    The generative model of our proposed method is calculated from the following equations:
    \begin{eqnarray}
    \pi &\sim &{\rm GEM}(\gamma), \label{pimm} \\ 
    z_{i} &\sim& {\rm Categorical}(z|\pi), \label{Z_imm}\\
    \Sigma_{k}^{-1} &\sim&  {\cal W}(\Lambda | V_{0},\nu_{0}), \label{sigmamm}\\
    \mu_{k} &\sim& {\cal N}(\mu | m_{0},(k_{0} \Lambda)^{-1}), \label{mu_kmm}\\
    x_{i} &\sim& {\cal N}(x| {\mu}_{z_{i}} , \Sigma_{z_{i}}),  \label{x_imm}\\
    C &\sim& {\rm Uniform}, \\ \label{Cmm}
    q_{s} &\sim& {\cal N}(q|\mu_{s},\Sigma_{s}) \label{q_smm1}, \\
    q_{s} &\sim& {\cal N}(q|\mu_{c_s},\Sigma_{c_s}), \label{q_smm2}
    \end{eqnarray}
    \delspan{where $\Lambda$ is a precision matrix calculated as an inverse matrix of a variance $\Sigma$. 
        ${\rm Categorical}(\cdot)$, ${\cal N}(\cdot)$, ${\rm Uniform}$,  and ${\cal W}(\cdot)$ are a categorical distribution, a multivariate Gaussian distribution, a uniform distribution, and a Wishart distribution, respectively.} 
        \addspan{where ${\rm Uniform}$ and ${\cal W}(\cdot)$ are a uniform distribution and a Wishart distribution, respectively.}
    The sampling processes in Equations (\ref{q_smm1}) and (\ref{q_smm2}) show that a Gaussian distribution corresponding to a child node $s$ and a Gaussian distribution of its parent node $c_s$ simultaneously draw the same data point $q_s$.
    \addspan{The latent variable $q_s$ represents joint points in the body map space. Meanwhile, $q_s$ is estimated by inference procedure of the DPGMM-LJ in a probabilistic manner.} 
    In the DPGMM-LJ, we assume a joint point as a latent data point $q_s$ simultaneously generated from two connected components, i.e., a joint point is a point that is included in two connected body parts. A joint point is drawn for each activated component $s_j \in \mathbf{S}=\{s_1, s_2, \ldots , s_S\} \subset \mathbb{N}$, where activated component means a cluster involving more than one sample.
    The tree structure $C$ is assumed to be drawn from a uniform distribution over trees that connect all of the activated component~\footnote{Note that the decision whether the $k$-th cluster is activated or not requires the information of $\{z_i\}_{\forall i}$. The relationship between  $\{z_i\}_{\forall i}$ and $\mathbf{S}$ is not described in the graphical model. However, this model simply ignores the effect for practical purpose. We assume that the $\mathbf{S}$ is determined by DPGMM's inference procedure.}.
    \addspan{The introduction of the simultaneous generation processes shown in Equations (\ref{q_smm1}) and (\ref{q_smm2}) is a crucial contribution. This simple modification can give the generative model to treat latent tree structure. In addition, simultaneous introduction of Gaussian distribution for emission distribution and uniform distribution for the prior distribution of the latent tree structure allow us to develop an efficient inference procedure, i.e., making use of the conventional minimum spanning tree algorithm as a part of its inference procedure.}
    \subsection{Inference of DPGMM-LJ}
    Our proposed method infers a body schema by using data points on a body map, which is constructed in advance.
    The body schema is inferred by a Gibbs sampling procedure, which is a type of Markov chain Monte Carlo (MCMC).
    Generally, Gibbs sampling is performed by using posterior distributions for each variable.
    Most of the posterior distributions of the latent variables except for $C$ can be derived easily because the model uses conjugate prior distributions.   
    
    The index of Gaussian distribution for the $i$-th data point is sampled from the following probability distribution:
    \begin{eqnarray}
    p(z_{i}|\pi,x_{i},\mu,\Sigma) \propto {\cal N}(x_{i}|\mu_{z_{i}},\Sigma_{z_{i}}){\rm Categorical}(z_{i}|\pi). \label{sample_Z_i}
    \end{eqnarray}
    The mean and variance of the $k$-th Gaussian distribution are sampled from the following probability distributions:
    \begin{align}
        p(\mu_{k}|\Sigma_{k},x,q,m_{0},k_{0},z,C) &\propto 
        \prod_{\forall i: z_i=k} {\cal N}(x_{i}|\mu_{k},\Sigma_{k}) 
        \prod_{\forall s : (s=k)\lor (c_s=k) }{\cal N}(q_s|\mu_{k},\Sigma_{k}) \nonumber\\ 
        &~~~\times{\cal N}(\mu | m_{0},(k_{0} \Lambda)^{-1} ), \label{sample_mu}\\
        p(\Sigma_{k}|V_{0},\nu_{0},k_{0},m_{0},\mu_{k},z,x,q,C)  
        &\propto 
        \prod_{\forall i: z_i=k} {\cal N}(x_{i}|\mu_{z_i},\Sigma_{z_i})\prod_{\forall s : (s=k)\lor (c_s=k) }{\cal N}(q_s|\mu_{k},\Sigma_{k}) \nonumber\\
        &~~~ \times  {\cal N}(\mu | m_{0},(k_{0} \Lambda)^{-1} ) {\cal W}(\Lambda | V_{0},\nu_{0}).
        \label{sample_sigma}
    \end{align}
    The position of a latent joint is sampled from the following distribution:
    \begin{eqnarray}
    p(q_{s}|C,\mu,\Sigma)  \propto   {\cal N}(q_s|\mu_{s},\Sigma_{s})
    {\cal N}(q_s|\mu_{c_s},\Sigma_{c_s}). \label{sample_q}
    \end{eqnarray}
    The posterior distribution of mixture weights of Gaussian distributions is sampled from the following distribution:
    \begin{eqnarray}
    p(\pi|\gamma,z) \propto  {\rm Dir}(\pi|\gamma) \prod_{\forall z_i \in z} {\rm Categorical}(z_{i}|\pi),
    \label{sample_pi}
    \end{eqnarray}
    where ${\rm Dir}(\cdot)$ is a Dirichlet distribution\footnote{Note that weak-limit approximation was conducted.}.

    However, calculating posterior distribution over possible $C$ results in a huge computational cost because the number of possible $C$ is exponentially large. Therefore, sampling $C$ from its posterior distribution is difficult to perform. On behalf of sampling $C$ from the posterior distribution, we update $C$ so as to maximize its posterior probability. 
    \begin{align}
        C^* &= \argmax_C P(C| \mu, \Sigma, q)\\
        &= \argmax_C  \sum_s \log P(c_s| \mu_s, \mu_{c_s}, \Sigma_s, \Sigma_{c_s} , q_s)\\
        &= \argmin_C  \sum_s  \underbrace{-\log \big( {\cal N}(q_s|\mu_s, \Sigma_s){\cal N}(q_s|\mu_{c_s}, \Sigma_{c_s}) \big)}_{\text{Quadratic cost function}} \label{eq:min_sp}
    \end{align}
    Note that the maximization problem of posterior probability can be reduced to the problem of finding a minimum spanning tree where each node corresponds to each Gaussian distribution.
    The minimum spanning tree is a tree that connects all the nodes, i.e., vertices, together with the minimal total weighting for its edges. In our settings, the weighting corresponds to the quadratic cost shown in Equation (\ref{eq:min_sp}).
    A conventional efficient minimum spanning tree search algorithm can be used to determine $C^*$.
    Our proposed inference method can reduce calculation cost by employing an efficient minimum spanning tree search algorithm.

    The body schema is updated using the minimum spanning tree search algorithm called Prim's algorithm~\cite{Cormen2001} at each iteration.
    The update of the body schema is performed by setting the optimized tree structure $C \leftarrow C^*$.
    The estimation of the latent tree structure and sampling of other latent variables are performed in an iterative manner. The total learning procedure is regarded as an approximate Gibbs sampling algorithm.
    This iterative learning process enables the DPGMM-LJ to estimate latent variables including latent tree structure $C$ from given data points.
    Note that the estimated tree structure $C$ can influence the position of a latent joint $q_{s}$ at the update phase in the next iteration and vice versa.
    Meanwhile, owing to the nature of the Bayesian nonparametrics, the number of clusters is also estimated in a probabilistic manner.  
    
    \section{Computational model for Body Schema Estimation}\label{sec4}
    Figure~\ref{fig:overview} shows the overview of the process with which our computational model estimates a body schema. The proposed method consists of the following two steps.
    First, a topological map is formed based on tactile information obtained from tactile sensors on the body, as shown at (a) in Figure~\ref{fig:overview}.
    The second step entails estimating a tree structure based on tactile information as shown at (b) in Figure~\ref{fig:overview}.
    
    \subsection{Body map generation}\label{subsec3-2} 
    To form a body map, we use a body map formation method obtained by modifying the method proposed by Olsson et al.~\cite{olsson2006unknown}. 
    The modified method uses the summation of conditional entropy for measuring the distance between two sensors. Meanwhile, the method uses multidimensional scaling (MDS) for placing the sensors appropriately in a low-dimensional space using the distance information~\cite{kruskal1964multidimensional}.
    Here, a body map is a model of a somatosensory map. 
    
    Olsson et al. defined the summation of conditional entropy as an information metric between sensors~\cite{olsson2006unknown}.
    The information metric $D$ satisfies the following metric space axioms: $D(X,Y)=D(Y,X){\rm \; \;(Symmetry)}$, $D(X,Y)= 0 {\rm \; \; iff \; \;} Y=X {\rm \; \;(Equivalence)}$, $D(X,Z) \leq  D(X,Y) +  D(Y,Z) {\rm \; \;(Triangle \;Inequality)}$.
    When the model has $M$ tactile sensors, the information metric $D(S_{i},S_{j})$ between two tactile sensors $i$ and $j$ is calculated by the equation
    \begin{eqnarray}
    D(S_{i},S_{j}) =H(S_{i}|S_{j})+H(S_{j}|S_{i})\label{end},
    \end{eqnarray}
    where $S_i$ and $S_j$ are random variables observed by two sensors $i$ and $j$, respectively.
    The conditional entropy $H(S_{i}|S_{j})$ is defined by the equation
    \begin{eqnarray}
    H(S_{j}|S_{i} )&:=&
    -\sum _{ x\in S_{i}  }{ \sum _{ y\in S_{j} }{ p_{ij}(x,y)\log _{ 2 }{ p_{ij}(y|x) }  }  }, \label{z} 
    \end{eqnarray}
    where $x$ and $y$ are tactile data and elements of $S_i$ and $S_j$, respectively.
    The joint distribution $p_{ij}(x,y)$ is calculated by using the following equation
    \begin{eqnarray}
    p_{ij}(x,y)&=& \frac{\sum _{ t=1}^{T}{ f(i,j,t,x,y)}}{T}. \label{xy212121} 
    \end{eqnarray}
    Here $t$ is an index number of tactile-information divided by a specified time interval $\Delta t$ and $T$ is a maximum index number of tactile-information data.
    $f(i,j,t,x,y)$ is defined by
    \begin{empheq}[left={  f(i,j,t,x,y)=  \empheqlbrace}]{align}
    1    \;  \;   \;& \; \; \; {\rm if}  \; \;  \;  a_{i,t}=x \land   a_{j,t}=y  ,  \nonumber  \\
    0     \; \;  \; & \;                      {\rm otherwise, }   \label{nnn3}
    \end{empheq}
    \addspan{where $i$ and $j$ are the sensor number, $t$ the specific time, and $x$ and $y$ are indices of quantized tactile information. The function $f(i, j, t, x, y)$ represents joint observation of two different sensors, $i$ and $j$.
        The memory $f(i,j,t,x,y)$ is used for calculating non-linear correlation between tactile sensors at each time step.
    }
    The tactile-information data $a_{i,t}$ and $a_{j,t}$ are quantized tactile information and have integer values between $1$ and $N$.
    The information metric thus defines the distance between two tactile sensors $i$ and $j$.
    Olsson et al. used an incremental algorithm to place sensors in the low-dimensional space referring to the measured distances between sensors. In contrast, for simplicity, we employ MDS to make an agent form a body map using the distance information calculated using the information metric. MDS is a widely used method of placing data in Euclidean space using distance information between data.
    Before running MDS, we need to determine the dimension $d$ of the low-dimensional space where the body map is formed.
    \subsection{Body schema estimation using DPGMM-LJ}
    Our proposed method estimates a body schema from a body map, which is formed by using the method described in Section~\ref{subsec3-2}. For estimating a body schema from the distribution of sensors in the low-dimensional space, i.e., body map, our computational model uses the DPGMM-LJ to estimate a body schema, i.e., a latent tree structure (see Section~\ref{sec3}).

    In summary, the overall procedure of our proposed computational model of body schema formation is described in Algorithm~\ref{alg:proc}.
    \addspan{In the Bayesian approach, the latent variables are attempted to be estimated as posterior distributions. Generally, MCMC provides samples from the posterior distribution. By gathering the samples, we can approximate the posterior distribution of the latent variables including the latent joint points. }
    
    Basically, there are two ways to use the results of the sampling procedure. One is to use the statistics of the sampled latent variables, i.e., Monte Carlo method. The other is to use the last sample of the drawn latent variables, i.e., point estimation.

    \renewcommand{\algorithmicrequire}{\textbf{Input:}}
    \renewcommand{\algorithmicensure}{\textbf{Output:}}
    \begin{algorithm}[tb]
        \caption{Overall procedure of body schema estimation}
        \label{alg:proc}
        \begin{algorithmic}
            \REQUIRE{$d$ //The dimension of the low-dimensional space in which a body map is formed }
            \STATE // General movements 
            \STATE Observe tactile information $\{a_{i,t}\}_{i=1,2, \ldots, M, t=1, 2, \ldots, T}$.
            \STATE // Body map formation
            \STATE Calculate the information metric ic $\mathbf{D} = \big( D(S_i,S_j) \big)_{\forall i,j}$ using $\{a_{i,t}\}_{\forall i,t}$.
            \STATE $\mathbf{x} \leftarrow {\rm MDS}(\mathbf{D} , d)$ // $\mathbf{x}$  A body map as data points in the $d$-dimensional space 
            \FOR{$k=1$ to {\rm MAX\_ITERATION}}
            \STATE Sample latent variables in DPGMM-LJ except for $C$ using Equations (\ref{sample_Z_i})--(\ref{sample_pi})
            \STATE Update $C$ using Equation (\ref{eq:min_sp})
            \IF{$k > {\rm BURN\_IN}$}
            \STATE Save latent variables
            \ENDIF
            \ENDFOR
            \ENSURE statistics of the saved latent variables
        \end{algorithmic}
    \end{algorithm}
    \section{Experiment 1: Body Schema Estimation}\label{sec5}
    To verify if the proposed computational model can estimate body schema using tactile information obtained through \addspan{random}\delspan{general} movements, the following experiment was performed in a simulation environment.
    Experimental results with different dimension $d$ of low-dimensional space for representing body map were also compared.
    \subsection{Condition}
\begin{figure}[tb]
        \begin{center}
            \subfigure[Exterior of the model]{
                \includegraphics[scale=0.15]{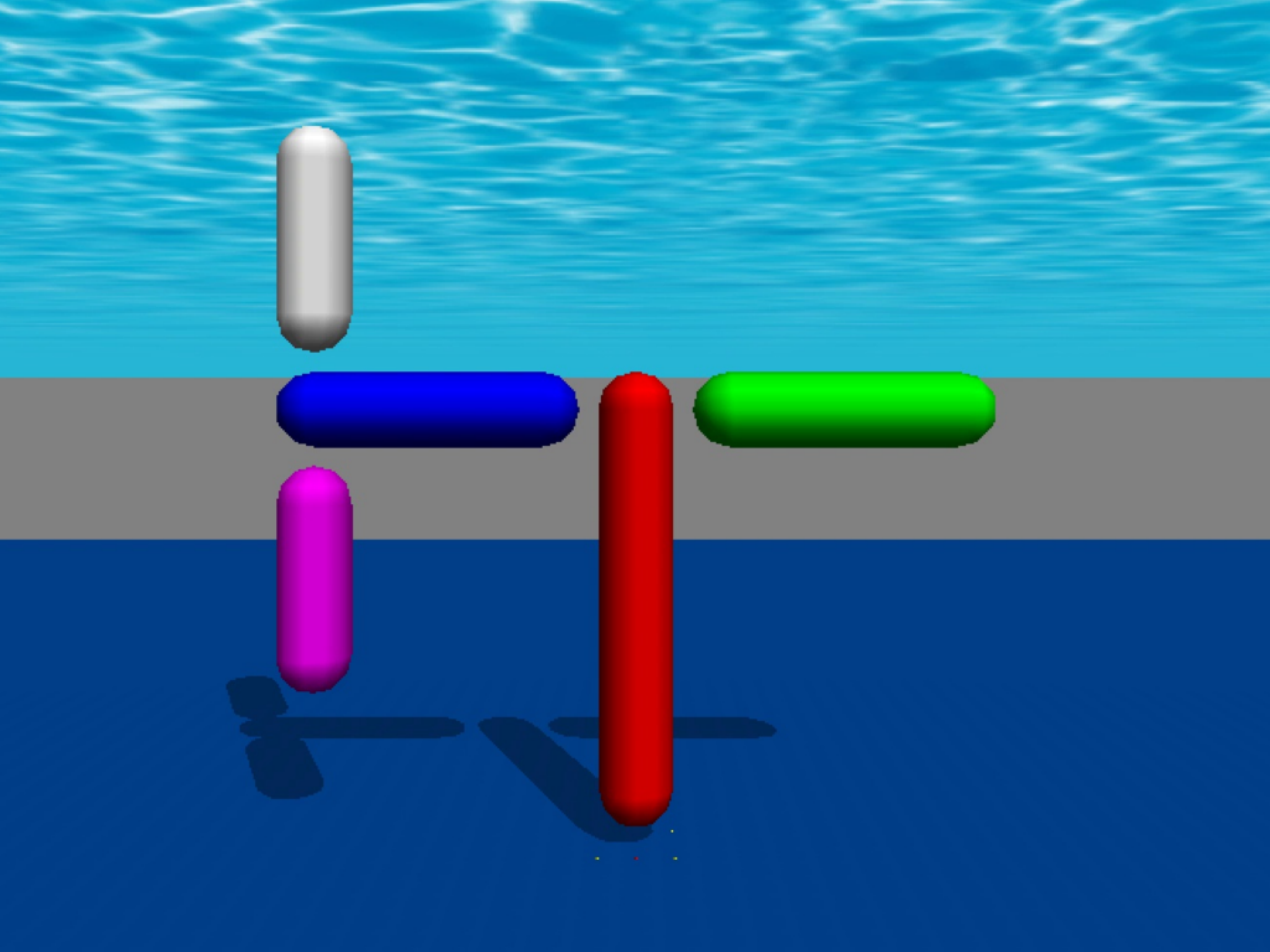}
            }
            \subfigure[Placement of sensors]{
                \mbox{\raisebox{-10mm}{\includegraphics[scale=0.42]{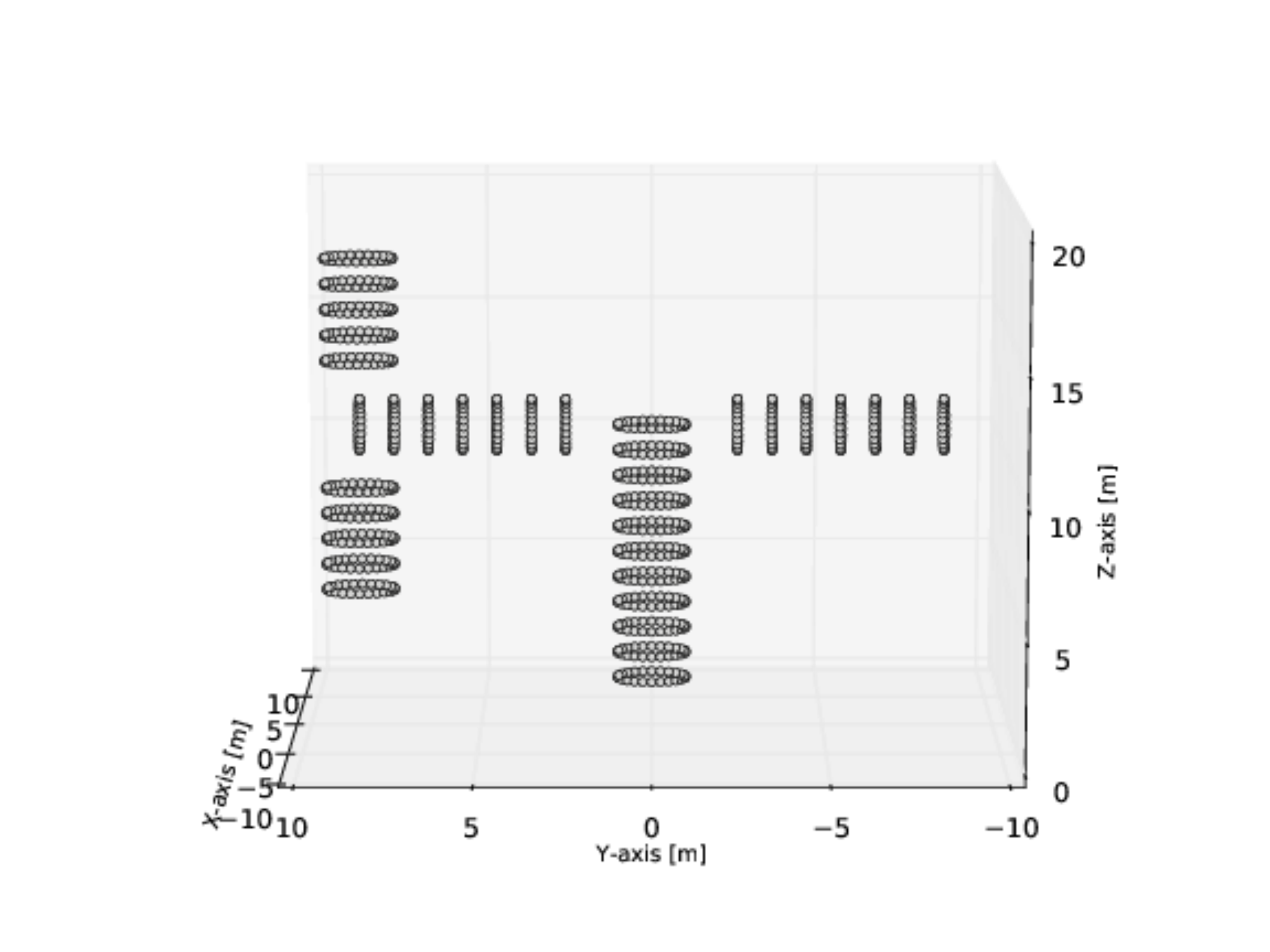}}}
            }
            \caption{Tree structure model for the verification experiments in virtual space
                \label{mimuratree}}
        \end{center}    
    \end{figure}

    The human body has a complex structure and multiple degrees of freedom.
    Therefore, preparing an accurate musculoskeletal model is itself a challenging problem. 
    Mori et al. dealt with this problem when they developed a constructive model of a fetus~\citep{Mori2010}.
    In this paper, we focus on the computational model. For the purpose, we kept the physical model as simple as possible.  
    A simple agent that has a tree structure instead of a precise human musculoskeletal model was prepared in a simulation environment, and used in the evaluation experiments (see Figure~\ref{mimuratree} (a)). The Open Dynamics Engine (ODE)~\cite{ODE} was used for simulating physical dynamics of the agent's movements in the evaluation experiments. 
    Each body part was represented by a cylinder, and each joint was modeled by a universal joint. All the physical environmental parameters that are not designated in this condition part were set to the default values of the ODE. 
    
    The target agent had five body parts as shown in Figure~\ref{mimuratree} (a). 
    On the surface of the five body parts, 840 virtual tactile sensors were attached (Figure~\ref{mimuratree} (b)).
    Mori et al.~\cite{Mori2010} took three factors, i.e., amniotic fluid, uterine wall, and self-contacts, into consideration.
    In this study, for simplicity, we only consider pressure of amniotic fluid.
    Each tactile sensor value was calculated using its three-dimensional position and direction at time intervals of $\Delta t = 0.1$ s. We assume that the liquid surrounding the agent does not have viscosity, and tactile information $f_{i,t}$ of sensor $i$ at time $t$ is measured by using the following equations:
    \begin{empheq}[left={  f_{i,t}=  \empheqlbrace}]{align}
    \rho \times(  \dot{p}_{i,t} \cdot n_{i,t})^2   \;  \;   \; \; \; \; & {\rm if } \;  0 <   \dot{p}_{i,t} \cdot n_{i,t},    \nonumber  \\
    0 \;  \;   \; \; \; \; \;  \;   \; \; \; \; & {\rm otherwise  } \label{b},
    \end{empheq}
    where $n_{i,t}$ is the normal vector to the surface of sensor $i$ at time $t$, $\rho$ is the density of amniotic fluid and is set to $1010 {\rm [kg/m^3]}$, and $\dot{p_{i,t}}$ is the speed of sensor $i$ at time $t$ and is calculated by using the equation 
    $\dot{p}_{i,t} = \frac{p_{i,t}-p_{i,t-1}}{\Delta t }\label{speed}$,
    where $p_{i,t}$ is the three-dimensional position of sensor $i$ at time $t$ and is obtained from the ODE. Note that the tactile-information data $a_{i,t}$ in Equation (\ref{b}) is a quantized value of $f_{i,t}$.
    The total time was 10000 [s] in the experiment.
    Thus the number of sampled tactile data, $T$, for each sensor was $100000$.
    The maximum number of clusters was set to $10$ for the weak-limit approximation.
    
\begin{figure}[tb]
        \begin{center}
            \includegraphics[width=1.0\linewidth]{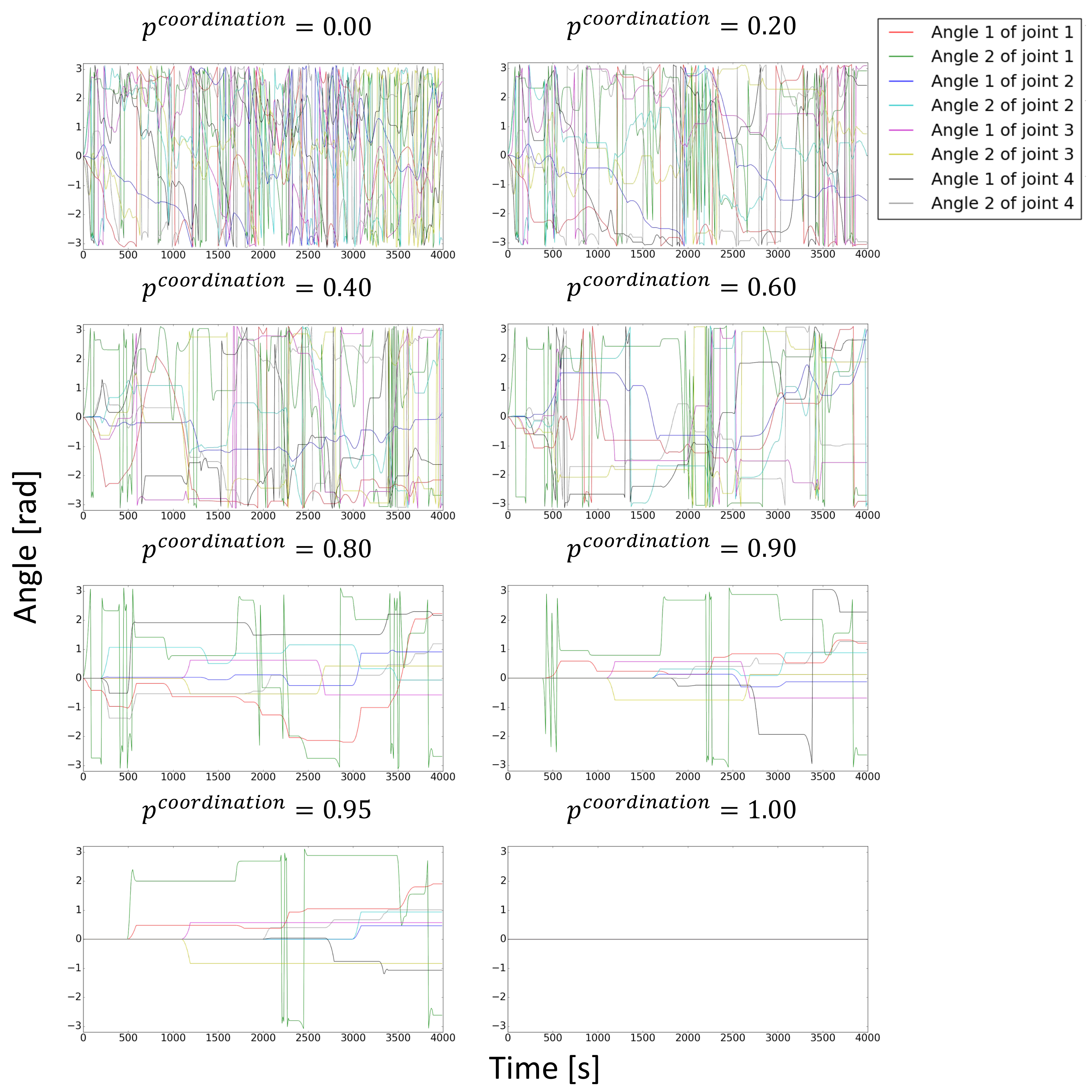}
            \caption{Time courses of joint angles in different $p^{\rm coordination}$ settings}
            \label{fig:Angle}
        \end{center}
    \end{figure}

    To model the structures embedded on random movements that is inspired by general movements in a simple way, we gave a simple model of motor coordination to the random movements of the agent. We made each joint move randomly with a specified probability $1-p^{\rm coordination}$.
    \addspan{    Each joint is locked at the beginning. At every 500 time step, a random variable is drawn for each joint and determines whether it starts moving with the probability $1-p^{\rm coordination}$. For each joint, the target angle is determined randomly by drawing a random variable from a uniform distribution over $[-\pi, \pi]$. In each step, actual torque input is calculated using a simple linear controller. The final directions and velocities of their movements were determined based on the physical simulation performed by ODE.  }
    In this experiment, the probability was set as $p^{\rm coordination} = 0.9$.
    The angle of each joint was fixed when it does not move. The parameter was determined empirically by observing the agent's movements. For example, the movements generated by $p^{\rm coordination} = 0.5$ looked lacking in structure that can be observed in a fetus's general movements. 
    We assume the probabilistic lock of the joints is a simple model of motor coordination.
    We call $p^{\rm coordination}$ the ratio of motor coordination in this study.
    \addspan{Figure~\ref{fig:Angle} shows the samples of time coursees of joint angles in different $p^{\rm coordination}$ settings.  }
    
    \subsection{Results}
    \begin{figure}[tb]
        \begin{center}
            \subfigure[Body map formed in two-dimensional space]{
                \includegraphics[scale=0.35,clip]{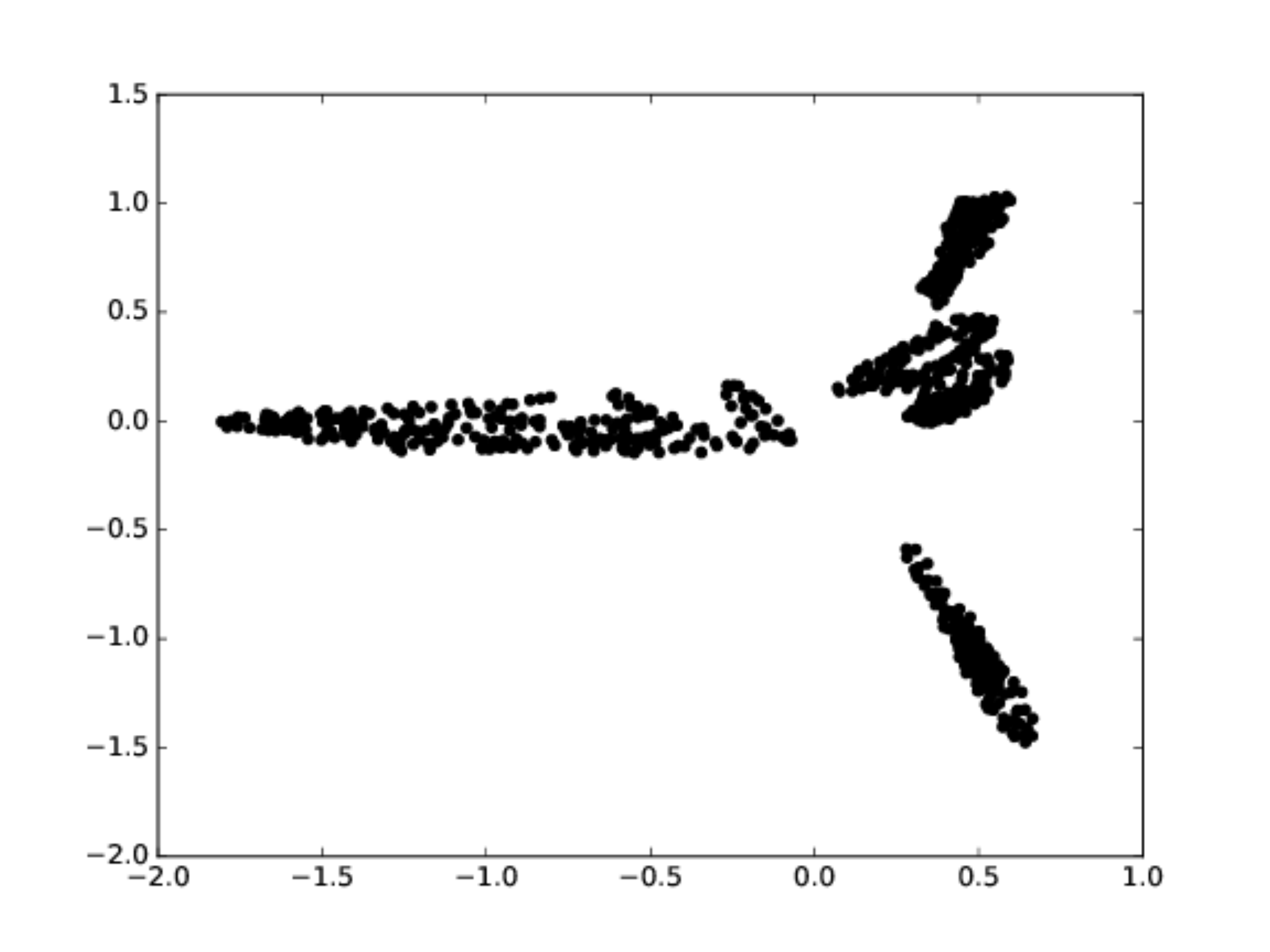}
            }
            \subfigure[Body map formed in three dimensional space]{
                \includegraphics[scale=0.35,clip]{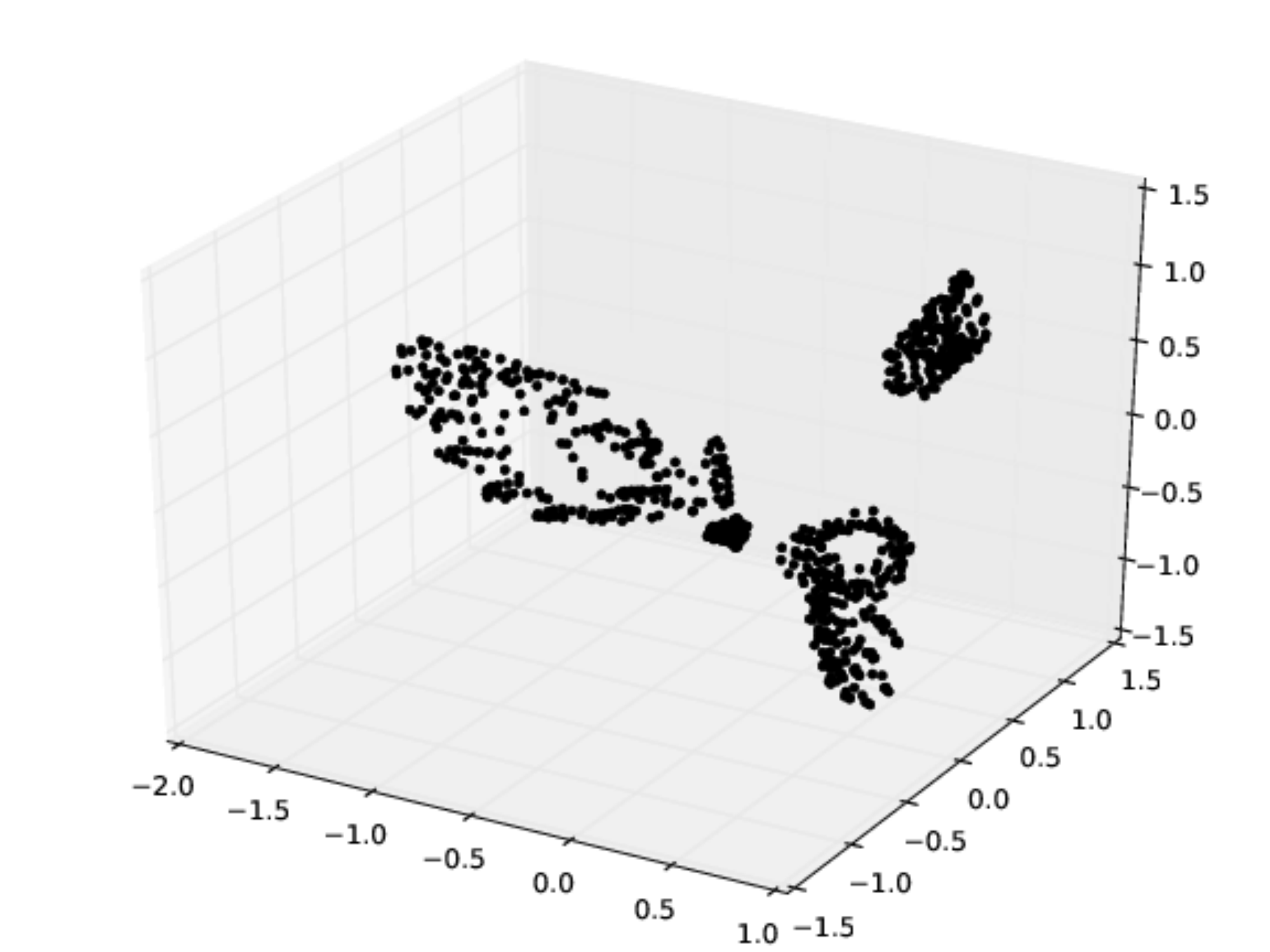}
            }\\
            \subfigure[The estimated clusters and joints from the 2D body map]{
                \includegraphics[scale=0.35,clip]{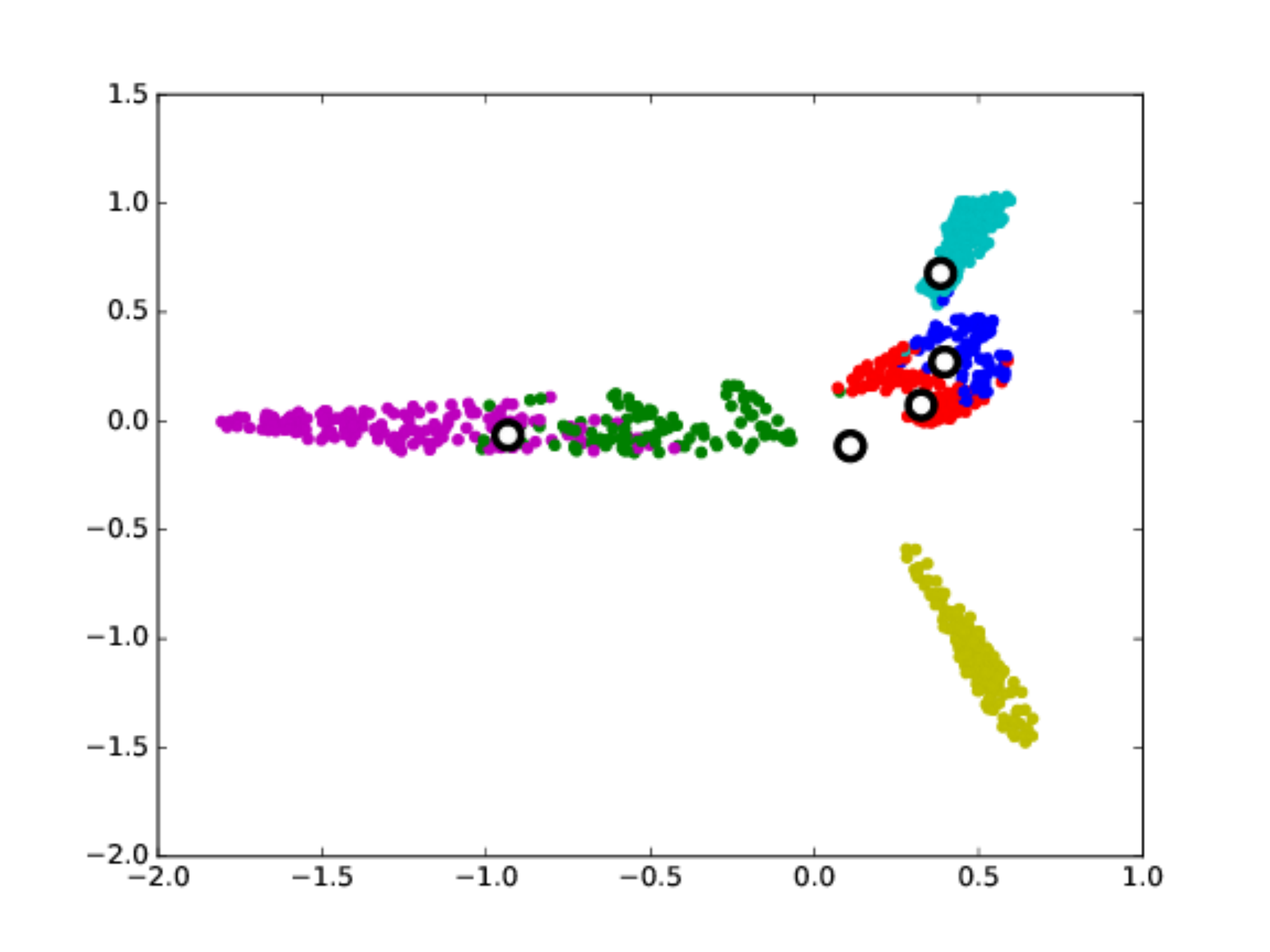}
            }
            \subfigure[The estimated clusters and joints from the 3D body map]{
                \includegraphics[scale=0.35,clip]{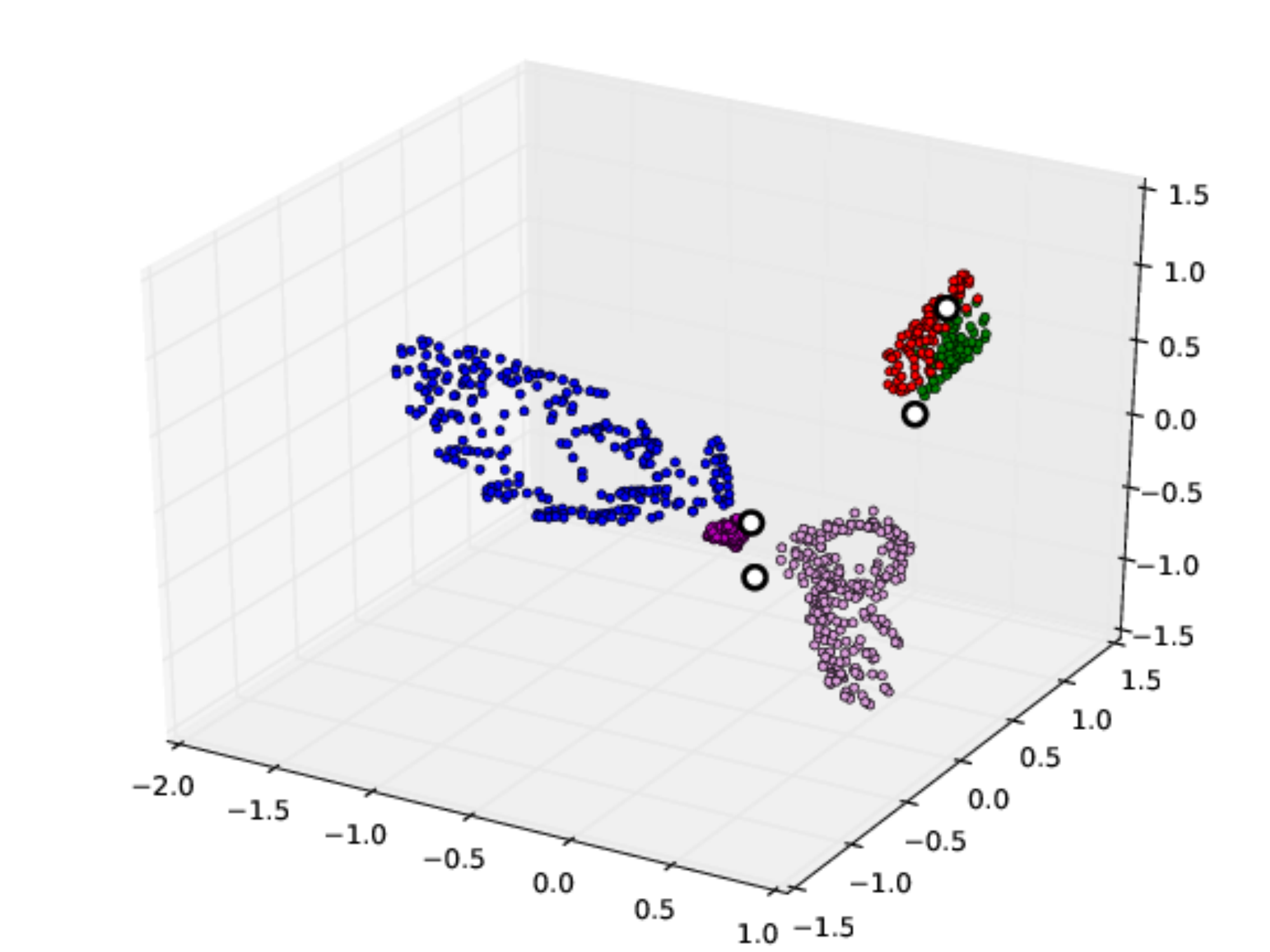}
            }
            \caption{Body map generation and joint estimation using the proposed method \label{body_map}}
        \end{center}    
    \end{figure}
    To investigate the effect of the capacity of the space for the body maps, we changed the dimension of the space from $d=2$ to $d=15$, and performed experiments to form
    the body schemata for each condition.
    Two samples of the estimation results are shown in Figure~\ref{body_map}. 
    Figure~\ref{body_map} (a) and (b) show body maps formed in the two- and three-dimensional spaces, respectively.
    Figure~\ref{body_map} (c) and (d) show samples of clustering results by DPGMM-LJ for $d=2$ and $d=3$. Different colors indicate different clusters, and the estimated joint points are plotted with small white circles from the 2D and 3D body maps respectively.
    \begin{figure}[tb]
        \begin{center}
            \subfigure{
                \resizebox*{15cm}{!}{\includegraphics{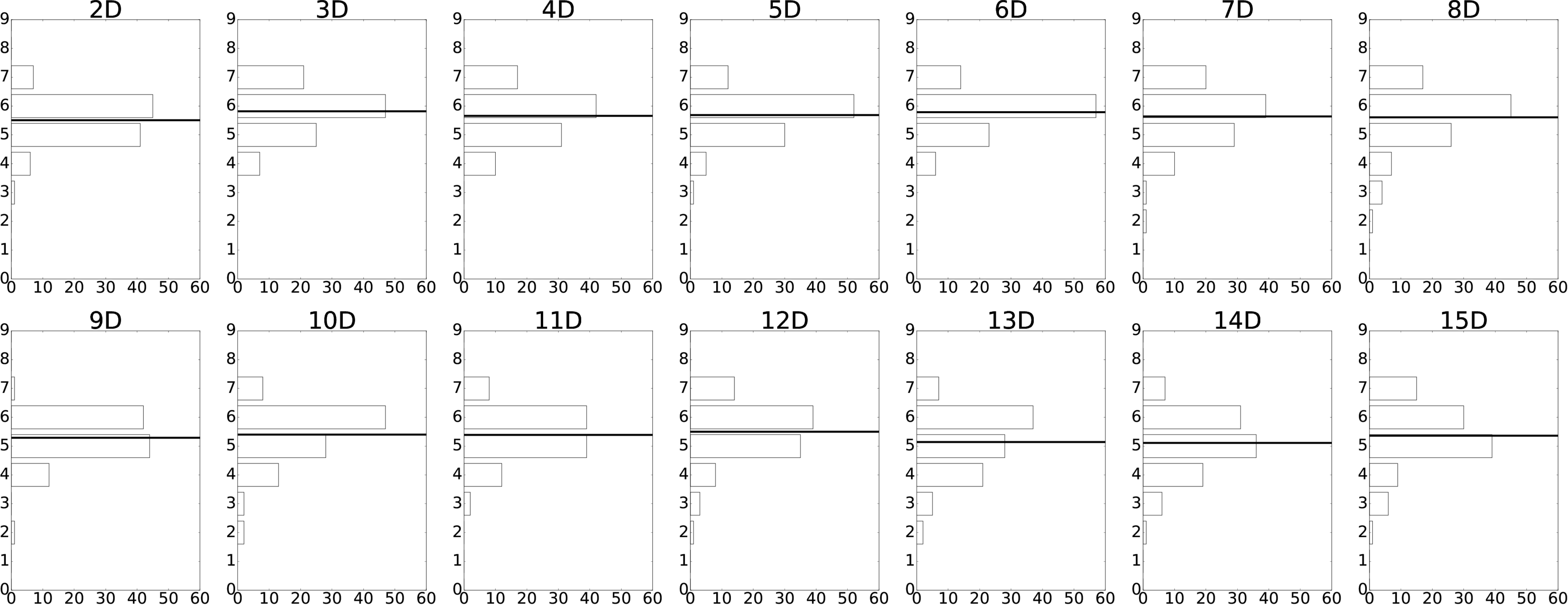}}}
            \caption{The number of estimated clusters in a body schema}
            \label{body_schema_nodes}
        \end{center}
    \end{figure}
    
    Quantitative evaluation of the clustering results, i.e., the estimation of the number of body parts and the correspondences between sensors and body parts, was performed.  Figure~\ref{body_schema_nodes} shows the number of clusters in a body schema estimated from formed body maps with multiple dimensions from $d=2$ to $d=15$ by using the proposed method.
    Each graph shows the histogram of the number of nodes sampled by the DPGMM-LJ, i.e., the estimated number of body parts, for each dimension with 100 samples.  
    In each figure, the vertical and the horizontal axes show the number of estimated clusters and the count of each number, respectively. The long horizontal lines show the average number of estimated nodes. In MCMC, the estimation results are represented by the samples from estimated posterior distributions and statistics calculated using the samples. 
    These figures demonstrated that our proposed model could estimate the number of clusters successfully even when the dimension of the space for a body map is $d=2$.
    The estimated number of the clusters was slightly larger than the correct number of body pars, i.e., $5$. The estimated number of the clusters means the number of clusters to which more than one data point was allocated. 
    Generally, a nonparametric Bayesian clustering method often forms some meaningless clusters, e.g., a cluster that involves only one data point. 
    Therefore, when we apply a nonparametric Bayesian clustering to a real dataset, 
    the number of estimated clusters tends to larger than its correct value.
    \addspan{    This owes to the assumption of Gaussian distribution for each cluster. The real data usually does not follow Gaussian distribution perfectly but have disturbed distributions and outliers. That tends to affect the clustering result so as to increase the number of clusters for modeling outliers. Even though the body structure of the agent in this simulation appears very artificial and simple, the task of body schema estimation is quite complex and challenging from the viewpoint of machine learning. }
    From this viewpoint, this result shows that the method could estimate the number of body parts \delspan{quite} appropriately. 
    \addspan{However, to overcome this problem, overestimation of the number of body parts, is, of course, our future challenge.}
    
    \begin{figure}[tb]
        \begin{center}
            \subfigure[Ward, k-means, and GMM]{
                \includegraphics[scale=0.12,clip]{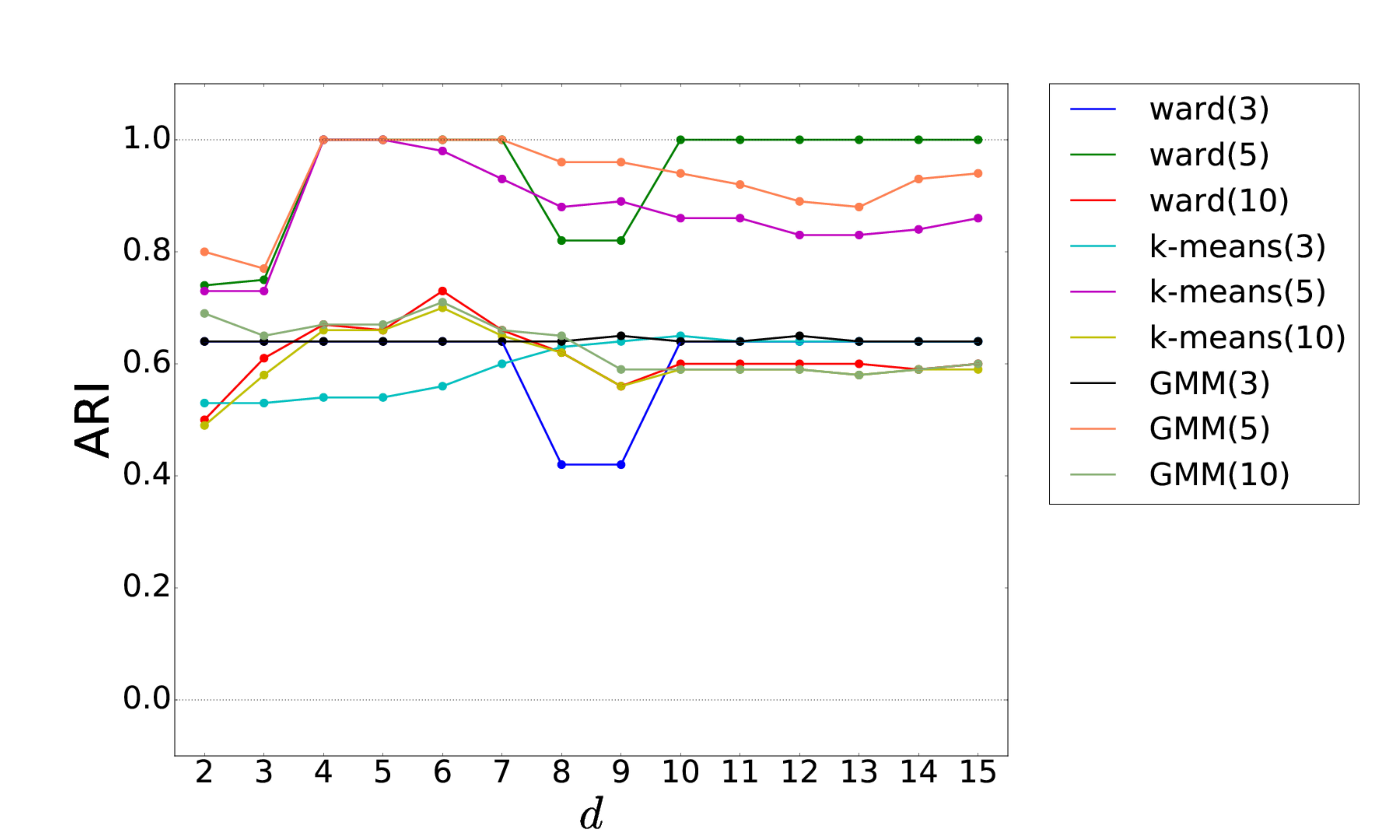}
            }
            \subfigure[DPGMM and DPGMM-LJ]{
                \includegraphics[scale=0.12,clip]{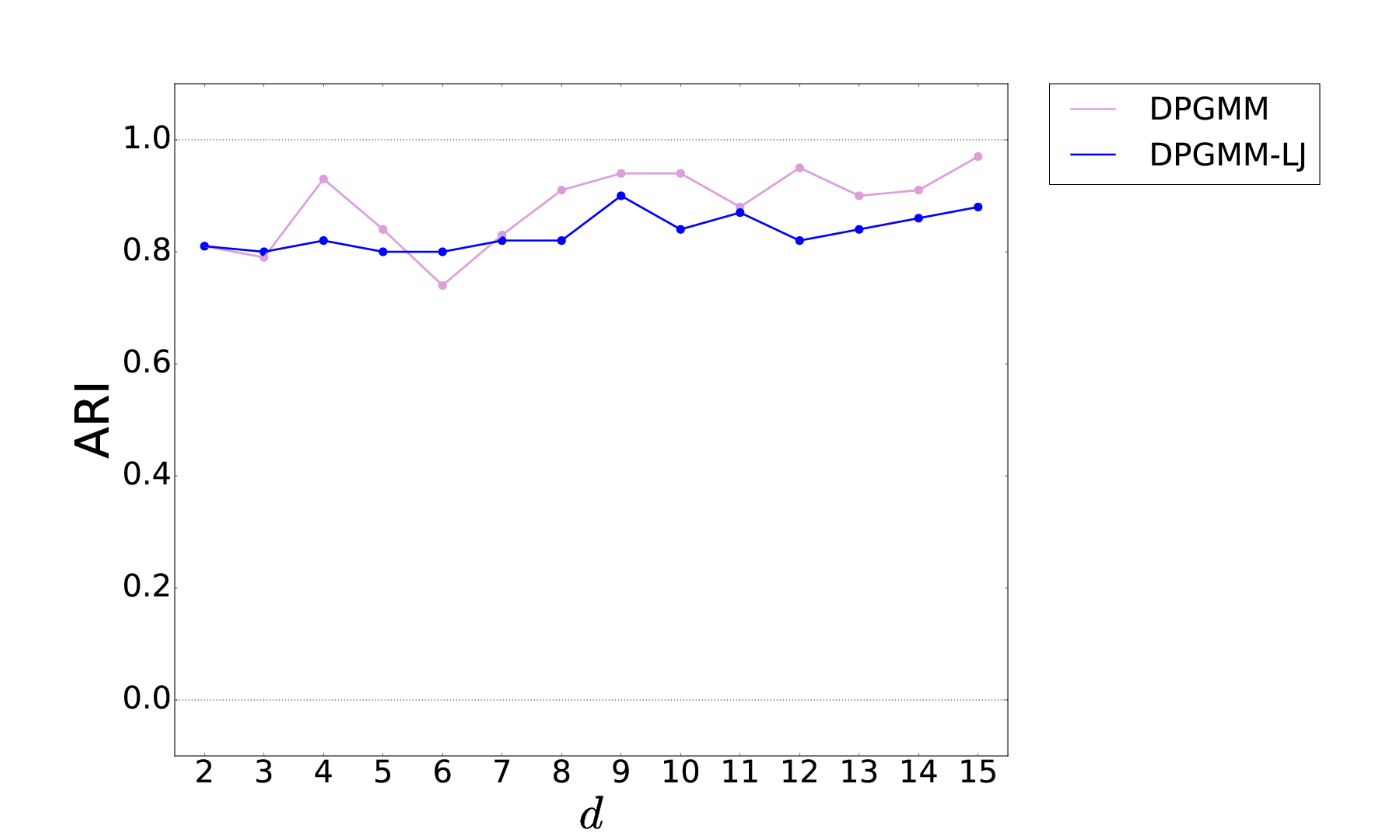}
            }
            \caption{The average ARI of clustering results in multiple dimensional spaces}
            \label{ARI_clustering}
        \end{center}    
    \end{figure}
    
    For evaluating clustering results, the adjusted Rand index (ARI)~\cite{hubert1985comparing} is used to measure the similarity between grand truth and a obtained clustering result in the evaluation experiments.
    If the similarity between ground truth and a clustering result is perfectly matched, the ARI becomes $1.0$. If the similarity is at chance level, the ARI is expected to be $0.0$.
    Figure~\ref{ARI_clustering} (a) and (b) show the average ARI of clustering results with different dimensions of the spaces where the body maps were formed. 
    We compared our proposed method with four different clustering methods, namely, Ward’s method, k-means, GMM, and DPGMM.  
    
    In each figure, the horizontal and the vertical axes show the dimensions and the ARI for each method, respectively. Among them, only the DPGMM can estimate the number of the clusters. 
    The number of clusters for Ward’s method, k-means, and GMM was set as 3, 5, and 10, respectively.
    First, Figure~\ref{ARI_clustering}~(a) shows that 
    Ward’s method, k-means, and GMM correctly estimated clusters with ARIs higher than 0.8 in most cases if the number of clusters was 5, which was the correct number of clusters. 
    However, when the number of clusters was set to incorrect values, i.e., 3 or 10, the ARIs became worse. 
    In contrast, even if the number of clusters was unknown, the DPGMM and the DPGMM-LJ correctly estimated clusters with ARIs higher than 0.8 as shown in Figure~\ref{ARI_clustering} (b).
    This result demonstrated that the DPGMM and the DPGMM-LJ, which are based on Bayesian nonparametrics, could estimate the number of clusters and cluster assignment simultaneously.
    However, no significant difference was observed between the results of the DPGMM and the DPGMM-LJ.
    As Figure~\ref{fig:gm} suggests, the difference between the DPGMM and the DPGMM-LJ is essentially small with regard to the clustering task. Note that the Markov blankets of $z_i$ in the two models are completely same, and only $\mu$ and $\Sigma$ are affected by $q$. 
    \begin{figure}[tb]
        \begin{center}
            \subfigure{
                \resizebox*{10cm}{!}{\includegraphics{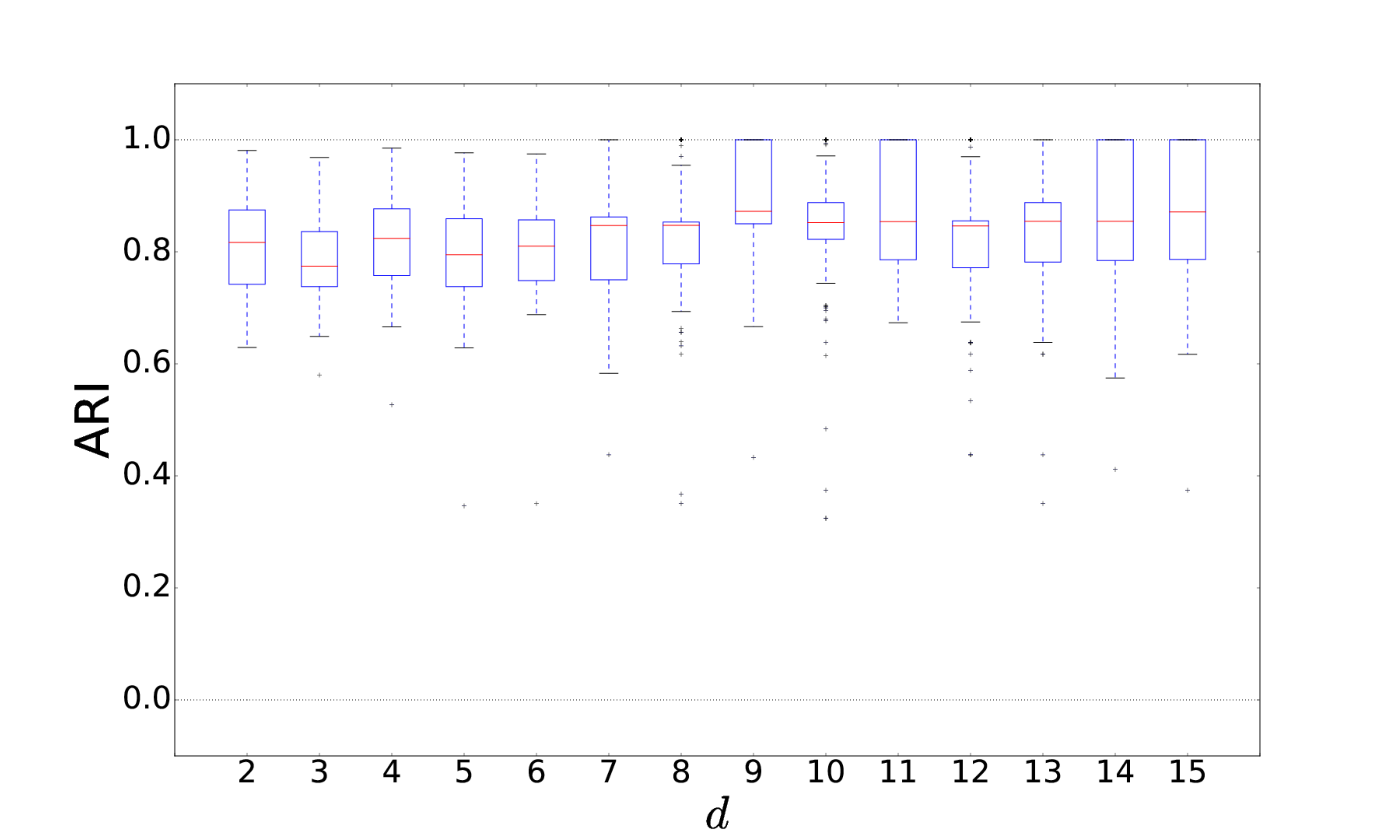}}}
            \caption{ARI of clustering results using DPGMM-LJ}
            \label{ARI_DPGMM-LJ}
        \end{center}
    \end{figure}
    
    Figure~\ref{ARI_DPGMM-LJ} shows the ARI of clustering results by the DPGMM-LJ with 100 samples using a box plot.
    In this figure, the horizontal and the vertical axes show dimensions and ARI of clustering results, respectively. Each blue box and a red line in the box show the dispersion and the median of the ARI.
    This figure demonstrated that the proposed method could appropriately estimate the clusters, i.e., the relationships between body parts and sensors, not only from the viewpoint of the average but also the distribution of the ARIs.
    \begin{figure}[tb]
        \begin{center}
            \subfigure[Comparative method (DPGMM + Prim's method)]{
                \includegraphics[scale=0.12]{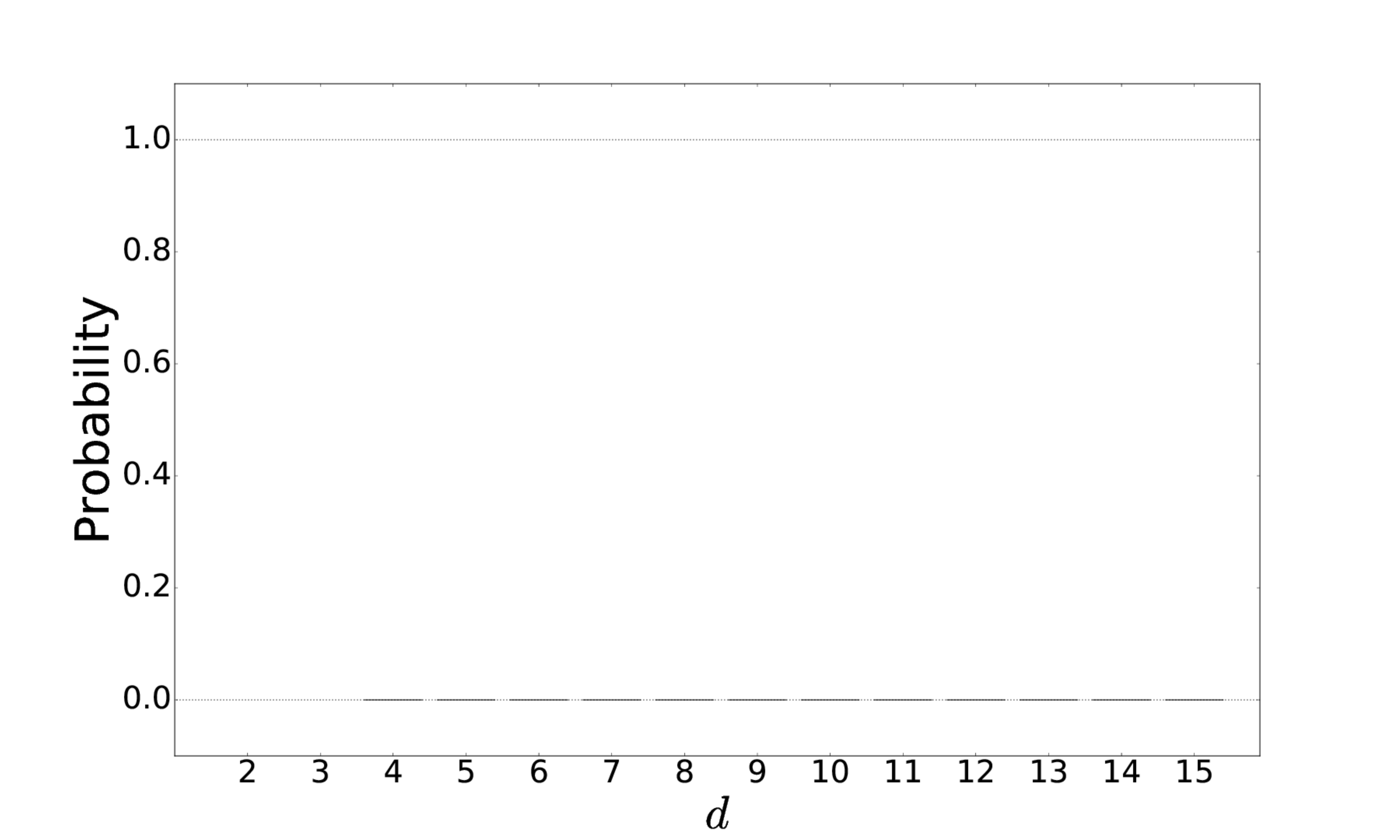}
            }
            \subfigure[DPGMM-LJ]{
                \includegraphics[scale=0.12]{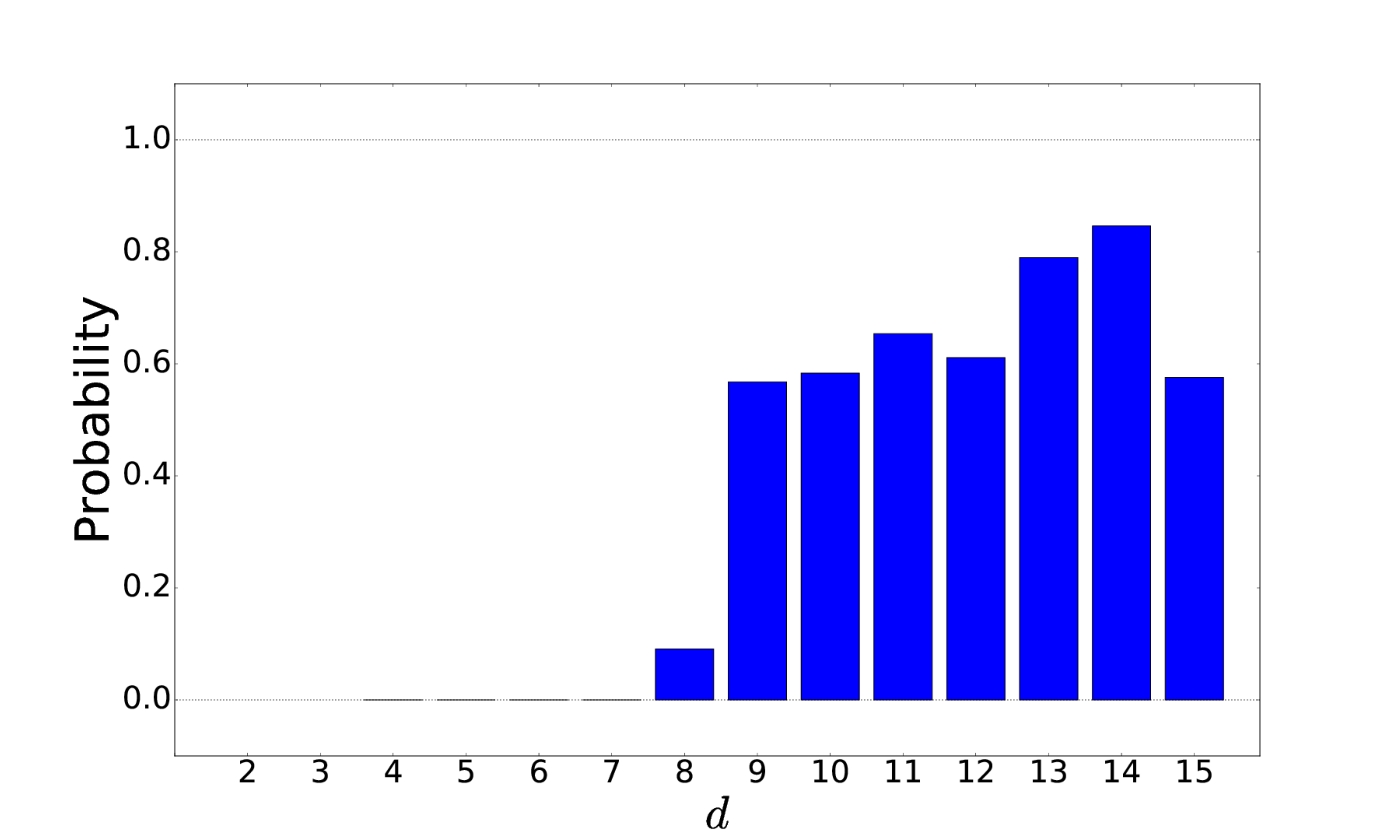}
            }
            \caption{Result of Experiment 1 for tree structure estimation}
            \label{Experimental1}
        \end{center}    
    \end{figure}
    
    To verify the effectiveness of the proposed computational model from the viewpoint of the estimation of kinematic structure, our proposed method, the DPGMM-LJ, was compared with a baseline method.
    The baseline method estimates clusters from the body map by using the DPGMM, and infer a latent tree structure by using Prim's algorithm based on the Euclidean distances between the centers of the estimated Gaussian distributions.
    Figure~\ref{Experimental1} (a) and (b) show the experimental results.
    In each graph, the horizontal and the vertical axes show dimensions and accuracy rate of the body schema estimation, respectively.
    As we mentioned above, the estimation results are obtained in a probabilistic manner. Inference results often contain small meaningless clusters. It is difficult to measure the similarity between probabilistic distribution over tree structures. Therefore, we focused on the samples that could estimate the number of body parts correctly and cluster the data points with ${\rm ARI} =1.00$. The success rates of tree structure estimation are calculated using the samples.
    
    As shown in Figure~\ref{Experimental1} (a), the comparative method was not able to estimate a correct tree structure in any trials. Note that the chance level of this estimation task is quite low.
    In contrast, as shown in Figure~\ref{Experimental1} (b), the DPGMM-LJ was able to estimate a correct tree structure with dimensions ranging from 9 to 15 with a relatively high probability.
    This result demonstrates that the proposed method can estimate the link structure of an agent solely from tactile information to some extent.
    The result also shows that a high dimensional space for a body map is required to infer an appropriate kinematic structure. 
    
    \section{Experiment 2: Body Schema Estimation with Different Ratio of Motor Coordination}~\label{sec6}
    Based on the positive results obtained in Experiment 1, we conducted the further investigation using our proposed constructive model. The importance of the motor coordination in general movements has been pointed out. However, its contribution to the body schema formation is not clear from the computational point of view.
    In this experiment, the performance of body schema estimation by the proposed computational model was evaluated with different ratios of motor coordination ($p^{coordination}$). The purpose of this experiment is to investigate the effect of motor coordination on body schema formation.
    \subsection{Condition}
    Most of the settings in the experiment are the same as those in Experiment 1 except the following. In this experiment, we changed the ratio of motor coordination. It was set as $p^{\rm coordination} \in \{0.00, 0.20, 0.40, 0.60, 0.80, 0.90, 0.95, 1.00\}$.
    The dimension of body map was set as 14 because the success ratio of body schema estimation was the highest in Experiment 1. 
    \subsection{Results}
    \begin{figure}[tb]
        \begin{center}
            \subfigure[Result of ARI using DPGMM-LJ]{
                \includegraphics[scale=0.12]{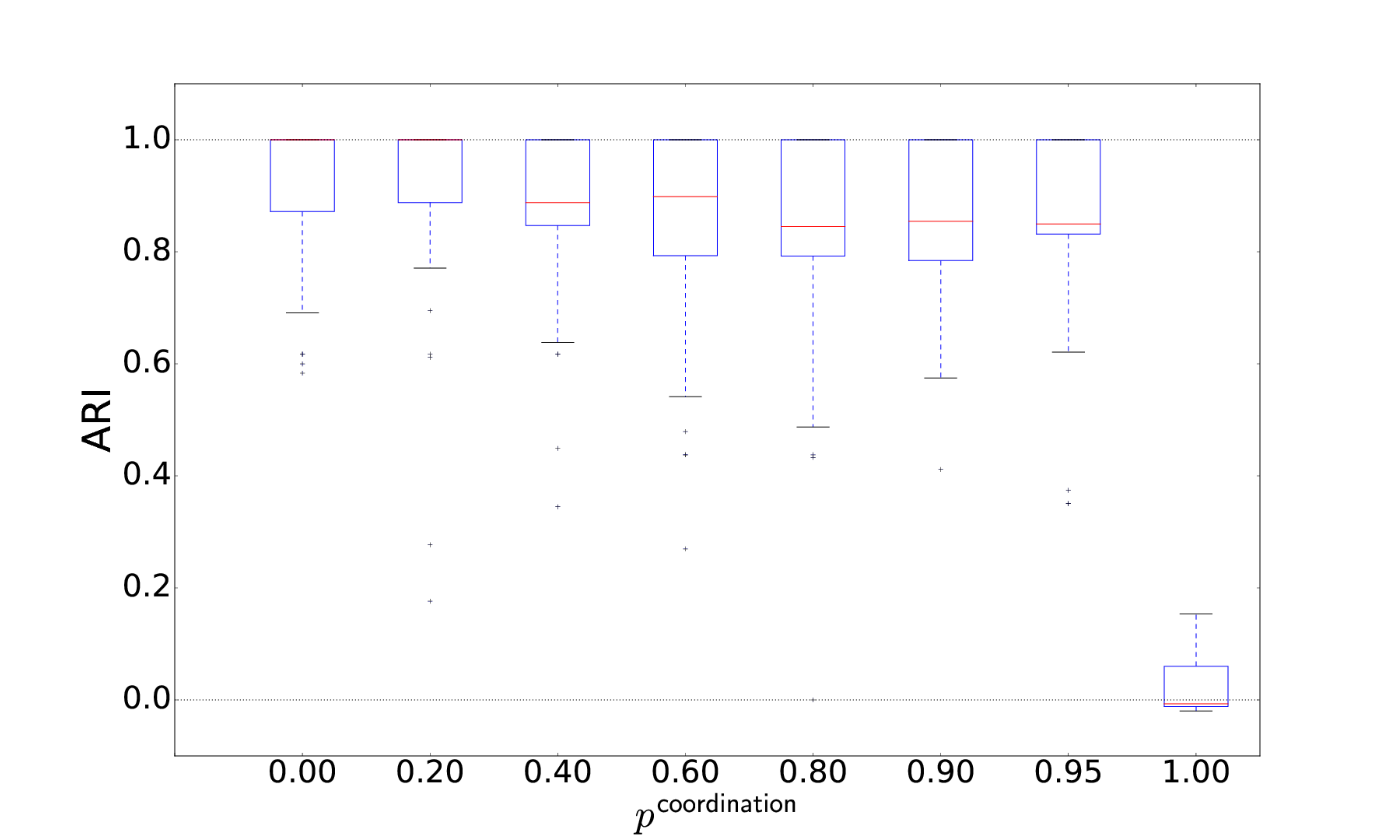}
            }
            \subfigure[Success rate of latent tree structure estimation]{
                \includegraphics[scale=0.12]{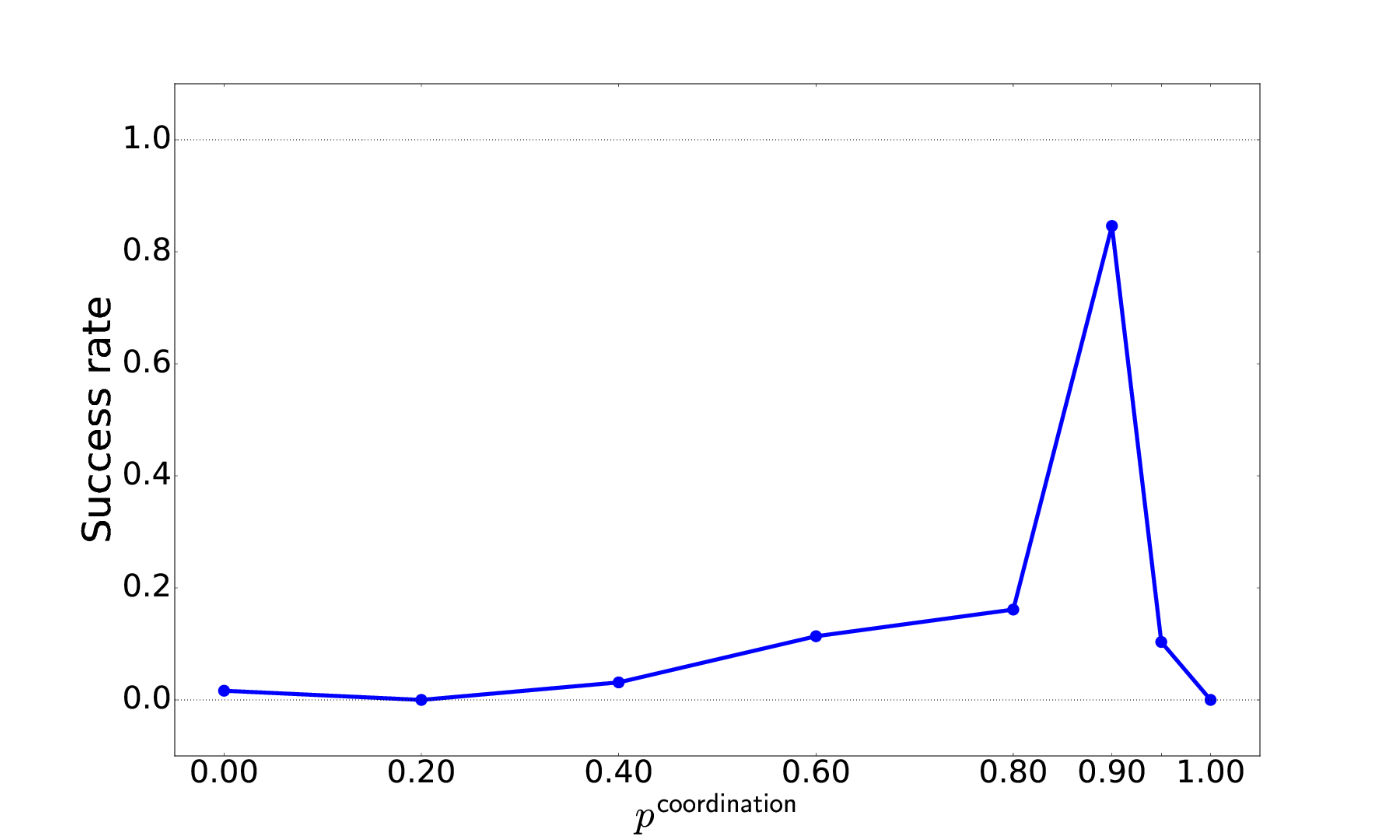}
            }
            \caption{Result of experimental 2 \label{Experimental2}}
        \end{center}    
    \end{figure}
    Figure~\ref{Experimental2} shows the experimental results of body schema estimation by our proposed computational model.
    Figure~\ref{Experimental2} (a) shows the ARI of clustering results obtained by using the DPGMM-LJ with different ratios of motor coordination.
    In the graph, the horizontal and the vertical axes show the ratio of motor coordination and the ARI, respectively.
    It is confirmed that the ARI was high regardless of the ratio of motor coordination except for the case $p^{\rm coordination} = 1.00$. When $p^{\rm coordination} = 1.00$, all the joints were always locked, and all body parts moved together. This means that the number of the body parts can be regarded as one. The result that the ARI is very low when $p^{\rm coordination} = 1.00$ is quite reasonable.
    
    In contrast, it is confirmed that the result of the tree structure estimation is clearly affected by the ratio of motor coordination $p^{\rm coordination}$. 
    Figure~\ref{Experimental2} (b) shows the success rate of body schema estimation by the DPGMM-LJ with different ratios of motor coordination.
    In the graph, the horizontal and the vertical axes show the ratio of motor coordination and the success rate of the tree structure estimation, respectively.
    In the same way as Experiment 1, we picked up the samples that could estimate the number of body parts correctly and classify the data points with ${\rm ARI} =1.00$, and calculated the success rate of tree structure estimation using them.
    It is confirmed that the accuracy rate for body schema estimation had the highest value $84.6 \%$ when the ratio of motor coordination is $p^{\rm coordination}= 0.90$.
    
    \subsection{Discussion}
    \addspan{In this section, we describe discussion about the two experiments. }
    
    \addspan{
        About body map formation, the results shown in Figure 6 are not so intuitive. It was observed that tactile sensors placed at the same body parts gathered together in the state space to a certain extent. However, the spatial relationships between body parts seemed not to be reconstructed in 2D and 3D space. Figure 10(b) shows that an approximately nine-dimensional body map space is required to reconstruct relationships between body parts, i.e., kinematic structure. It suggests that we can find that the body map formation method can reconstruct the relationships between body parts in the high dimensional space if we can visualize the high-dimensional space.}
    
    \addspan{
        The proposed method can simultaneously infer clustering results and a latent tree structure. It can be expected that such simultaneous optimization could increase the clustering performance as well. However,  Figure 8 (b) showed that there was no significant difference between the two methods. The DPGMM-LJ even performed worse than DPGMM in many conditions. In DPGMM-LJ each cluster forcibly related and connected.  It is considered that the mutual constraints decreased clustering performance of DPGMM-LJ.}
    
    \addspan{Consider the influence of structure embedded in the agent's random motion.}
    At first glance, \addspan{Figure 11 (b)}, the range of $p^{\rm coordination}$ where the DPGMM-LJ could estimate an appropriate tree structure looks narrow. However, it does not indicate a disadvantage of our proposed computational model, but clearly suggests the importance of motor coordination in body schema formation in the case of the constructive approach.
    
    It is known that a fetus's general movements involve a certain degree of motor coordination.  
    Quantifying the degree of motor coordination of a fetus and mapping onto the computational model appropriately are difficult tasks.  
    Note that the \addspan{}general movements performed by $p^{\rm coordination} = 0.90$ looked more natural than other settings even though the evaluation is qualitative and even subjective.
    The result suggests that the motor coordination in \addspan{random}\delspan{general} movements has an important contribution to the body schema formation. Further investigation is required. 
    
    The computational explanation of this phenomenon is as follows. While the estimation of the number of body parts requires statistical differences between two sensors included by two different body parts, the estimation of the latent tree structure requires statistical similarity between two sensors placed on two connected body parts.
    Completely random movements tended to enhance the independence of the statistical distribution of tactile information by two sensors placed on two different body parts. The higher motor coordination gave, the higher dependency between two sensors placed on two connected body parts. However, too high coordination of body parts, i.e., stiffed movements, can harm the body schema formation as shown in Figure~\ref{Experimental2} (b) ($p^{\rm coordination} = 1.00$).  This means that the well-balanced coordination between body parts is important to form an appropriate body schema. 
    
    
    \section{Conclusion}~\label{sec7}
    In this paper, we proposed the DPGMM-LJ and a computational model that can estimate the body schema of an artificial agent in a simulation environment solely from its tactile information. We regarded a body schema as a kinematic structure that can be represented by a tree structure.  
    The proposed model involving randomly moving multilink body system in a simulation environment and the computational model for body schema estimation is a constructive model of a fetus in a womb that conducts general movements.
    The computational model consisted of two elements: body map formation method inspired by Olsson et al.~\cite{olsson2006unknown}, and the DPGMM-LJ.
    A simulation experiment showed that the computational model could estimate the body schema of the artificial agent correctly in the simulation environment.
    This also showed that the DPGMM-LJ could determine appropriate clusters, the number of clusters, joint points, and latent tree structures. 
    The result of another experiment suggested that enough amount of motor coordination, which is represented by the rate of locked joints in the experiment, in general movements, and the sufficient dimension of representing the body schema are important to determine the body schema correctly, even though estimating the number of body parts does not require such conditions. The fact that general movements are not totally random movements, but structured movements is gathering attention. Our simulation experiment in this paper provides a new result suggesting that such a structure plays an important role in body schema formation.
    
    We have developed a novel constructive model that can reproduce the body schema formation process of a fetus in mother's womb.
    Insofar as we know, the proposed model is the first computational model that can estimate a body schema, i.e., latent tree structure, solely from the tactile information obtained through \addspan{random}\delspan{general} movements. 
    The proposed computational model is a strong candidate for a model that explains the body schema formation process because no other computational model can achieve the same task, i.e., body schema estimation from \addspan{random}\delspan{general} movements, currently. Future challenges are as follows. 
    
    \addspan{
        In the experiment, the body structure of the agent has the minimum complexity to test our proposed method, DPGMM-LJ. Performing experiments using a more precise and complex fetus model to test our method is desirable. However, creating a realistic fetus model itself is a very challenging research topic. Mori et al. have been conducting studies related to this topic~\cite{Mori2010}. To perform experiments on a more realistic fetus model and to evaluate our method is one of our future challenges. 
    }
    
\delspan{Phantom limbs pain is a representative example of disorders prompted by abnormality in body schema in the brain.
    Phantom limbs pain is caused by the difference between one's body schema and his/her substantial body. This means that the disorder is caused by the inadaptability of the patient's body schema. 
    The phantom limb pain is considered to be caused by the difference between substantial body and its body schema. To discuss phantom limbs from the computational viewpoint, it is desirable to have a computational model that can explain the dynamics of body schema including changes in the kinematic structure. Now, we have such a computational model. 
    To model phantom limbs by extending the proposed constructive model is our next challenge.}  
    
    Meanwhile, body schema should be formed not only using tactile information but also using visual and motor information. 
    However, our current model does not take visual information into consideration. 
    \addspan{For example, Chang et al.~\cite{Chang2016} used a similar approach to extract kinematic structure from visual information even though they used nonparametric approach and did not used nonparametric Bayesian approach. By following a similar approach, we consider that our method, DPGMM-LJ, will be able to be applied to visual information as well. After the extension, to make DPGMM-LJ have multimodal emission distributions including visual and tactile information will be the next step.} Extending our model to integrate visual and motor information for body schema formation is one of our future challenges.
    
    In addition, the current computational model consists of two separate modules, i.e., body map and body schema formation. However, such processes could be mutually related in our brain. As we observed in the experimental results, the performance of the tree structure estimation is still limited.
    To develop a more sophisticated model that can optimize a body map and a body schema simultaneously is also our future challenge. The integration of the tactile sensory information and visual information will also contribute to the improvement of the performance of the body schema estimation process.
    \section*{Acknowledgment}
    This research was partially supported by a Grant-in-Aid for Scientific Research on Innovative Areas 2015-2017 (15H01670) funded by the Ministry of Education, Culture, Sports, Science, and Technology, Japan.
    \bibliographystyle{tADR}
    \bibliography{mimura}
\end{document}